\documentclass[preprint,12pt]{elsarticle}

\journal{Information Systems}

\usepackage{amssymb}
\usepackage{amsmath}
\usepackage[defblank]{paralist}

\usepackage[algo2e,lined]{algorithm2e}
\usepackage{algorithm}

\usepackage{xspace}
\usepackage{xcolor}
\usepackage{subfigmat}
\usepackage[capitalise]{cleveref}
\usepackage{url}

\newcommand{\declare}{\textsc{Declare}\xspace}

\newcommand{\constraint}[1]{\textsf{#1}}
\newcommand{\activity}[1]{\textnormal{\textsf{#1}}}

\newcommand{\tasks}{\Sigma}
\newcommand{\alltasks}[1]{#1.\tasks}
\newcommand{\ctasks}[1]{#1.\tasks^c}

\newcommand{\somehps}{\mathcal{S}}
\newcommand{\proc}{P}

\newcommand{\lang}[1]{\mathcal{L}(#1)}

\newcommand{\trace}{\rho}

\newcommand{\proj}[2]{#1|_{#2}}

\newcommand{\tup}[1]{\langle #1 \rangle}
\newcommand{\set}[1]{\{ #1 \}}

\newtheorem{definition}{Definition}

\begin{document}

\begin{frontmatter}

\title{On the Hybrid Nature of ABPMS Process Frames \\and its Implications on Automated Process Discovery}

\author[inst1,inst2]{Anti Alman}
\ead{anti.alman@ut.ee, anti.alman@unibz.it}
\author[inst3]{Izack Cohen}
\ead{izack.cohen@biu.ac.il}
\author[inst4]{Avigdor Gal}
\ead{avigal@technion.ac.il}
\author[inst2]{Fabrizio Maria Maggi}
\ead{maggi@inf.unibz.it}
\author[inst2]{Marco Montali}
\ead{montali@inf.unibz.it}

\affiliation[inst1]{
            organization={University of Tartu},
            addressline={Narva mnt 18}, 
            city={Tartu},
            postcode={51009},
            country={Estonia}
        }
\affiliation[inst2]{
            organization={Free University of Bozen-Bolzano},
            addressline={via Bruno Buozzi 1}, 
            city={Bolzano},
            postcode={39100},
            country={Italy}
        }
\affiliation[inst3]{
            organization={Bar-Ilan University},
            city={Ramat Gan},
            postcode={5290002},
            country={Israel}
        }
\affiliation[inst4]{
            organization={Technion – Israel Institute of Technology},
            city={Haifa},
            postcode={3200003},
            country={Israel}
        }

\begin{abstract}
A core component of any AI-Augmented Business Process Management System (ABPMS) is the \emph{process frame}, which gives the system process-awareness and defines its maximal behavioral boundaries. Compared to traditional process models, the process frame should, in principle, provide a somewhat more permissive representation of the managed processes, such that the (semi) autonomous behavior of an ABPMS, referred to as \emph{framed autonomy}, could emerge. In addition, the process frame is not limited to a single linguistic or symbolic formalism and may incorporate heterogeneous knowledge ranging from predefined procedures to common sense rules and best practices. In this paper, we first conceptualize the ABPMS process frame as a hybrid business process representation, consisting of semi-concurrently executed procedural and declarative process models, extending the open-world assumption of the declarative paradigm also to procedural models. The latter allows any set of (non-conflicting) models of either type to be combined for execution, but complicates the automated discovery of these models from event data. Existing approaches for procedural models are particularly affected due to their reliance on observing directly-follows relations between pairs of activities. In search of an alternative, we present an in-depth analysis of how different procedural behaviors manifest as sets of discovered \declare constraints, each corresponding to a specific type of eventually-follows relation. This reveals behavioral overlaps between declarative and procedural models, while also laying the foundation for developing corresponding process (frame) discovery techniques.
\end{abstract}


\end{frontmatter}

\section{Introduction}
\label{sec:introduction}

The research manifesto on AI-Augmented Business Process Management Systems (ABPMSs) outlines a vision in which the traditional BPMS lifecycle is augmented with AI technologies, enabling the system to “reason about the current state of the process (or across several processes) to determine a course of action that improves the performance of the process” \cite{DBLP:journals/tmis/DumasFLMMRACGFGRVW23}. A core component of any ABPMS is the \emph{process frame}, which provides the system with process awareness and defines the boundaries in which the system must operate to achieve its goals. The latter represents a fundamental shift from a system that supports the execution of business processes, potentially through automating some tasks, towards a system that exhibits a degree of autonomy over how the processes are executed, which decisions are taken therein, and what optimizations are implemented over time, with the end-goal of uplifting business process management systems to a level of autonomous decision-making comparable to human actors.

In this paper, we first propose an approach for representing the process frame of an ABPMS based on the general guidelines provided in~\cite{DBLP:journals/tmis/DumasFLMMRACGFGRVW23}. We take inspiration from hybrid business process representations~\cite{DBLP:journals/is/AndaloussiBSKW20} and the recent call for holistic business process management~\cite{DBLP:conf/bpm/BandaraLRM21}, combining both with our earlier works on the enactment, monitoring, and discovery of interacting business processes~\cite{DBLP:journals/is/AcitelliAMM25,DBLP:conf/pmai/AlmanCGMM24,ALMAN2023102271,ALMAN2023102512,DBLP:conf/bpm/WittlingerAAMM25} and on developing the multi-model paradigm~\cite{DBLP:conf/caise/AlmanMRRW24,Maggi2026}. In particular, we conceptualize the notion of an ABPMS process frame as a hybrid business process representation, consisting of semi-concurrently executed procedural and declarative models, moving beyond a single modeling formalism and supporting the incorporation of heterogeneous knowledge. In this context, we argue for adopting the open-world assumption of the declarative modeling paradigm also for procedural models, such that each model of the process frame, declarative or procedural, will impose explicit execution requirements and/or limitations only on the activities within that model. This enables language-agnostic combining of any (non-conflicting) models of either type, formally defined through trace projections, but comes with significant implications on various process mining tasks.

Chief among these is the task of automated discovery, more specifically, the discovery of individual procedural models as compositional semi-concurrent components of the overall process frame in the presence declarative process constraints. We highlight fundamental limitations of most procedural process discovery approaches in this setting and explore the feasibility of overcoming these limitations by relying on declarative process discovery also for procedural models. As in our earlier works, we utilize Petri nets~\cite{DBLP:journals/pieee/Murata89,DBLP:journals/topnoc/HeeSW13a} and \declare~\cite{DBLP:conf/edoc/PesicSA07} as representative languages for procedural and declarative process models, respectively, with deterministic finite state automata (DFAs), as defined in~\cite{DBLP:conf/bpm/Westergaard11} specifically, used as the ``bridge'' between the declarative and procedural paradigms. This results in the second contribution of this paper --- an in-depth analysis of how common procedural behaviors manifest through declarative constraints, laying the foundation for developing of corresponding process (frame) discovery techniques.

The rest of this paper is structured as follows. \cref{sec:frame} presents our conceptualization of the ABPMS process frame, including the corresponding formalization, and outlines the resulting implications on process discovery. \cref{sec:behaviors} analyses the manifestation of procedural behaviors through declarative constraints, under the open-world assumption, including minor extensions of the \declare modeling language towards accurately detecting most common procedural behaviors. \cref{sec:related} describes the initial research trends related to ABPMS, followed by an overview of other related works. Finally, \cref{sec:conclusion} concludes the paper by outlining future research directions.
\section{The ABPMS Process Frame}
\label{sec:frame}

\begin{figure}[t]
    \centering
    \includegraphics[width=\textwidth]{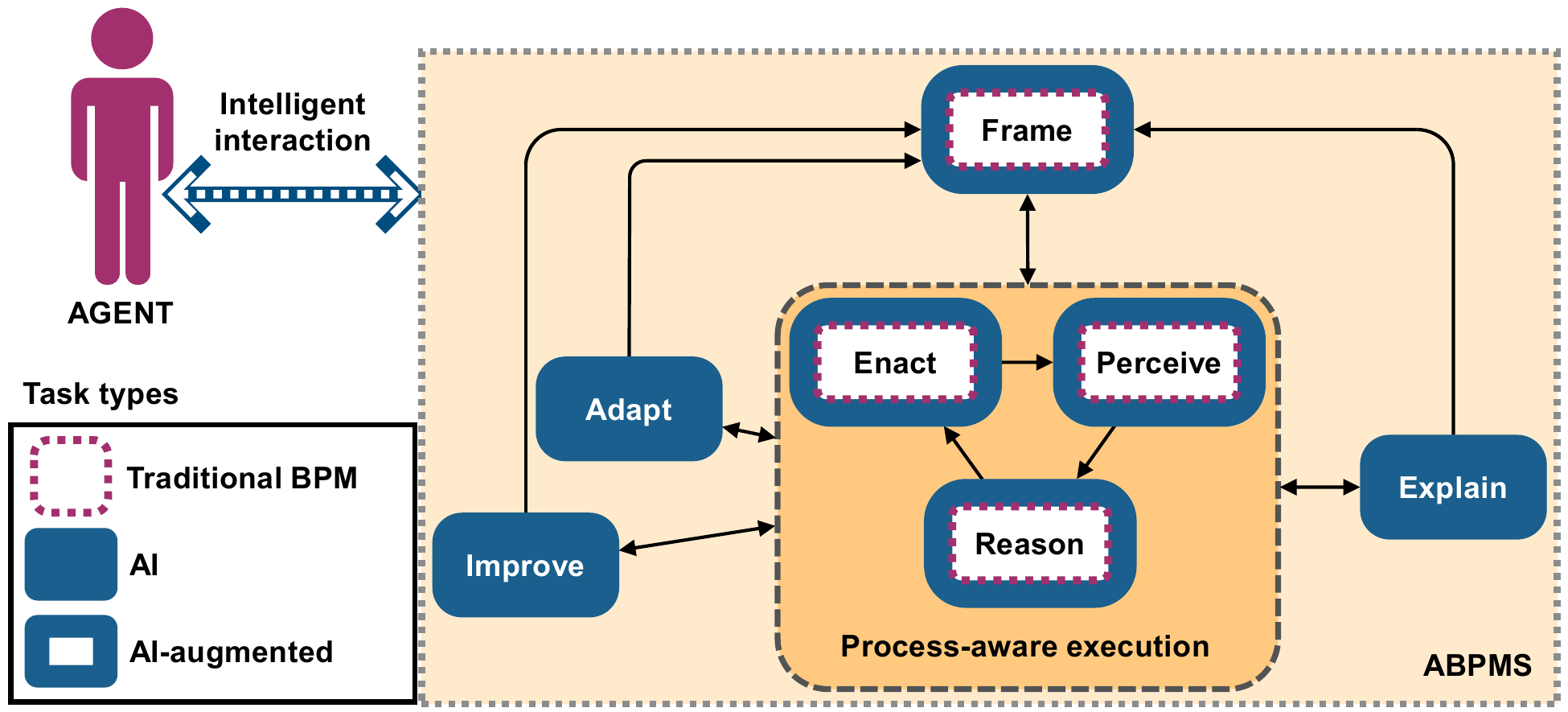}
    \caption{ABPMS lifecycle (taken from~\cite{DBLP:journals/tmis/DumasFLMMRACGFGRVW23}). Rounded rectangles represent lifecycle steps, while arrows denote sequencing.}
    \label{fig:abpms_lifecycle}
\end{figure}

The operational lifecycle of an ABPMS (cf. \cref{fig:abpms_lifecycle}) involves two actors: the ABPMS itself and one or more (not necessarily human) agents interacting with the ABPMS. Classifying the ABPMS, as one of the actors, is a fundamental difference from traditional business process management systems, reflecting that the ABPMS should have \emph{autonomy} over how the process (or processes) are executed. The exact level of autonomy can vary, but it should not be absolute. Instead, the boundaries applying to human actors, in terms of how the process(es) can be executed and modified, should also apply to the autonomy of the ABPMS. This may include, for example, legal requirements (e.g., obtaining a permit before starting construction), internal operating procedures (e.g., mandatory review of management decisions), organizational strategy (e.g., maintaining production costs below some threshold), and other similar knowledge that governs or is otherwise relevant for the optimal execution of the managed process(es).

The question of how to express such knowledge, as a process frame, is left open in~\cite{DBLP:journals/tmis/DumasFLMMRACGFGRVW23}, with procedural and declarative process modeling languages, including their hybrid combinations, being suggested as possible options. Reusing existing process modeling languages, especially ones with precise execution semantics, is indeed a natural starting point, given the central role of process-aware execution in the ABPMS vision (cf. \cref{fig:abpms_lifecycle}). However, this entails a conceptual shift from explicit modeling of complete process executions to defining the behavioral boundaries of an ABPMS based on heterogeneous sources of process knowledge, which, in turn, warrants a corresponding paradigm shift in how process modeling languages are used.

In the following, \cref{sec:conceptualization} outlines our conceptualization of the ABPMS process frame as an amalgamation of existing modeling languages, \cref{sec:formalization} provides the corresponding formalization, \cref{sec:example} describes a minimalistic example, and \cref{sec:discovery_implications} highlights the resulting implications on automated process discovery.

\subsection{Conceptualization}
\label{sec:conceptualization}

Contrary to traditional process models, procedural ones in particular, a process frame is not required to provide a complete specification of an entire business process.\footnote{In this paper, we often use the term ``specification'' instead of the term ``model'' to further stress this difference.} Instead, the actual executed process(es) should emerge, within the boundaries of the process frame, through the reasoning capabilities of the ABPMS itself. To support this, the notion of a process frame, as described in~\cite{DBLP:journals/tmis/DumasFLMMRACGFGRVW23}, is broader than the notion of a traditional process model. In particular, a process frame:
\begin{inparaenum}[\it (i)]
    \item is not limited to a single linguistic or symbolic formalism;
    \item may incorporate heterogeneous knowledge ranging from predefined procedures to common sense rules and best practices; and
    \item should, in principle, provide a more permissive process representation, such that the (semi) autonomous behavior of the ABPMS could emerge.
\end{inparaenum}

We propose reusing existing process modeling languages for creating the process frame, as also suggested in~\cite{DBLP:journals/tmis/DumasFLMMRACGFGRVW23}, but with a layer of overarching interpretation in support of the aforementioned broader characteristics. In that direction, we take inspiration from hybrid business process representations~\cite{DBLP:journals/is/AndaloussiBSKW20} and the recent call for holistic business process management~\cite{DBLP:conf/bpm/BandaraLRM21}, combining both with our earlier works on the enactment, monitoring and discovery of interacting business processes~\cite{DBLP:journals/is/AcitelliAMM25,DBLP:conf/pmai/AlmanCGMM24,ALMAN2023102271,ALMAN2023102512,DBLP:conf/bpm/WittlingerAAMM25} and with our proposal for developing and adopting the multi-model paradigm for business process management~\cite{DBLP:conf/caise/AlmanMRRW24,Maggi2026}.

Similar to~\cite{DBLP:journals/tmis/DumasFLMMRACGFGRVW23}, we do not fix the concrete modeling languages at the conceptual level. However, a distinction is made between procedural and declarative specifications, following the corresponding process modeling paradigms, where procedural refers to definitions of concrete activity sequences, and declarative refers to definitions of temporal activity relations and activity cardinalities over complete execution sequences. For example, a procedural specification may require \activity{A}, \activity{B}, \activity{C} to be executed in that exact sequence, while a declarative specification may require every \activity{A} to be eventually followed by \activity{D} at some, otherwise unspecified, future point in time.

\begin{figure}[t]
    \centering
    \includegraphics[width=\textwidth]{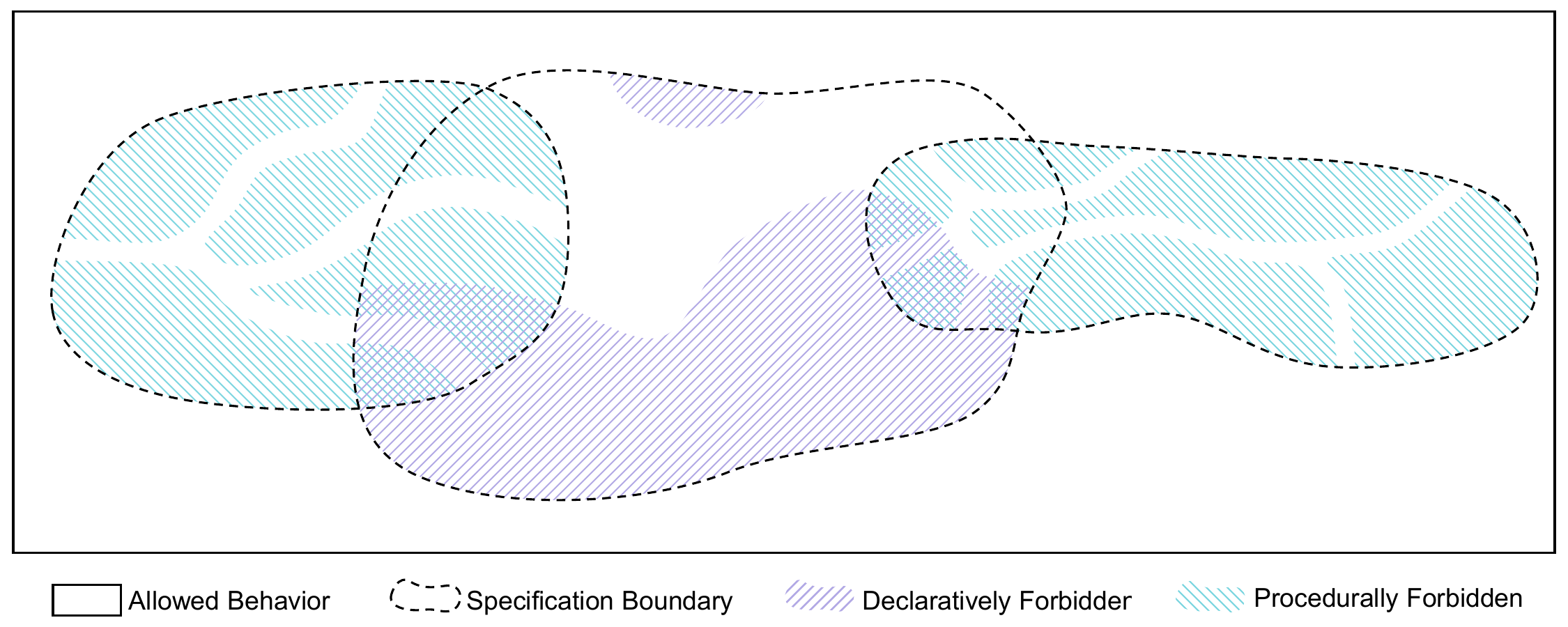}
    \caption{Conceptual view of a multi-model ABPMS process frame.}
    \label{fig:paradigm_concept}
\end{figure}

The starting point of our conceptualization is the open-world assumption of the declarative modeling paradigm, i.e., every activity execution, that does not contradict a given model, is considered to be allowed by that model~\cite{thesis_Pesic}. This alone introduces inherent permissiveness into the process frame, but it also forms the foundation for moving beyond a single modeling formalism and supporting the incorporation of heterogeneous knowledge. More specifically, it enables compositionality of individual process specifications, akin to the compositionality of constraints in the \declare language~\cite{DBLP:conf/bpm/AlmanCMMA21}, such that the existence of one specification does not preclude the existence of another, but rather, serves as a further refinement of the overall allowed behavior. The general idea is to think of each individual specification as some logical component of the overall behavioral boundary of the ABPMS, whether that component is a process, a sub-process, a common procedure, a dependency between some activities, or an orchestration of processes, sub-processes, or procedures, etc. Such specifications will then interact through behavioral overlaps, which, in this paper, correspond to sub-sequences of activity executions being governed by multiple specifications.

A conceptual view of the resulting process frame is given in \cref{fig:paradigm_concept}. The open-world assumption is adopted for individual specifications, such that any behavior outside some specification boundary is left open from the perspective of that specification, but may still be covered by a different specification.\footnote{This contradicts with the usual ``closed-world'' interpretation of procedural process models, where all behavior is assumed to be represented. For example, alignment approaches, such as the one in~\cite{DBLP:conf/bpm/BarWL25}, usually penalize the execution of activities outside the model as ``move in log only''.}
The specification boundaries may overlap, in which case, the allowed behavior corresponds to the intersection of the behaviors that are allowed by all specifications within the given overlap. Meanwhile, the union of specification boundaries corresponds to all behaviors that the process frame is aware of, i.e., there may be other potentially relevant behaviors outside the scope of the current process frame. If the ABPMS can observe such behaviors, then it should be decided if these behaviors are implicitly allowed or forbidden, or the process frame should be expanded by modifying the individual specifications or by adding new ones, potentially as part of autonomous reframing within the boundaries of a corresponding meta-frame~\cite{DBLP:journals/tmis/DumasFLMMRACGFGRVW23}.

The concrete implementation of this type of process frame, also referred to as a \emph{multi-model process frame}, will depend on the specific modeling languages being used. For example, one may use process modeling languages, such as, BPMN~\cite{BPMN_spec}, Petri nets~\cite{DBLP:journals/pieee/Murata89,DBLP:journals/topnoc/HeeSW13a}, \declare~\cite{DBLP:conf/edoc/PesicSA07}, DCR Graphs~\cite{DBLP:conf/bpm/DeboisHS14,DBLP:journals/corr/abs-1110-4161}, but also business rule modeling languages, such as, RuleML~\cite{DBLP:conf/semweb/BoleyTW01} or SBVR~\cite{SBVR_spec}, and potentially even ontology languages, such as, OWL~\cite{mcguinness2004owl} or OntoUML~\cite{DBLP:conf/iceis/BenevidesG09}, or languages from the realm of agent oriented modeling~\cite{sterling2009art}. However, note that combining any of these languages, at least for behaviors in terms of activity executions, requires adopting the open-world assumption, if not already adopted by the language, and defining how the intersection of allowed behaviors is determined. The former effectively means that a specification should, in general, ignore any activity executions not included in that specification, while the latter can be solved in a largely language-agnostic manner, as shown in \cref{sec:formalization}.

\subsection{Formalization}
\label{sec:formalization}

We start with some preliminary notions. Given a finite set of activities $\tasks$, a trace $\trace$ over $\tasks$ is a finite sequence of activities in $\tasks$. To clearly distinguish the activities contained in a trace, we use the tuple notation, e.g., the activity sequence \activity{a}, \activity{b}, \activity{c} is denoted by $\tup{a,b,c}$, and an empty trace is denoted by $\tup{}$. As customary, we denote the set of traces over $\tasks$ as $\tasks^*$. Since the process frame, as described in \cref{sec:conceptualization}, contains multiple ``local'' specifications operating over overlapping, distinct sets of activities, we need to define what such specifications see of a ``global'' trace defined over them. We do so through the notion of \emph{trace projection}. Given a trace $\trace \in \tasks^*$ and a subset $\tasks' \subseteq \Sigma$ of activities, the projection of $\trace$ on $\tasks'$, written $\proj{\trace}{\tasks'}$ is the sequence obtained from $\trace$ by removing all elements that are not in $\tasks'$. 

In the most general sense, a \emph{process specification} defined over a set $\tasks$ is nothing else than a subset of $\tasks^*$, identifying which traces belong to (or conform with) the specification (or, equivalently, which traces the specification accepts). As the set of traces belonging to a process specification may be infinite, one uses a compact, finite description (such as, an accepting Petri net or a set of \declare constraints). We generically assume that, for such a compact description~$\proc$, we can then characterize the language of $\proc$, written $\lang{\proc}$, as the set of traces in $\tasks^*$ captured by $\proc$.

\begin{definition}[Process specification]
\label{def:ps}
A \emph{process specification} is a pair $\tup{\tasks,\proc}$, where $\tasks$ is a finite set of activities and $\proc$ is a process description over $\tasks$. $\tup{\tasks,\proc}$ accepts trace $\trace \in \tasks^*$ if $\trace \in \lang{\proc}$.
\end{definition}

We formalize a \emph{process frame} as a composition of process specifications defined over overlapping sets of activities. Such specifications can be seen as \emph{local} process specifications that \emph{synchronize} on their \emph{shared activities}. This is substantiated as follows --- whenever an activity $t$ is performed, local specifications that do not have $t$ among their own activities will simply ignore it, while those that have it will all execute it. This implicitly requires that such specifications will all need to be in a state where $t$ is enabled, and will update their internal state by executing $t$ concurrently.

Two observations are in place. First, we are not dictating \emph{how} such local specifications are internally structured, i.e., they may be specified using Petri nets (with or without silent transitions and repeated labels), or other temporal/dynamic formalisms.
Second, \emph{global constraints} applied to all specifications can still be defined by introducing a specification that sees the whole set of activities of other specifications. The interplay across specifications can then be captured by requiring that, given a trace over the entire set of activities, every specification (including the one containing global constraints) accepts the projection of the trace on its own set of activities.

\begin{definition}[Process frame]
\label{def:hps}
A \emph{process frame} $\somehps$ is a finite set of process specifications $\tup{\tasks_1,\proc_1}, \ldots, \tup{\tasks_n,\proc_n}$, called \emph{specifications}. The \emph{set of  activities} of $\somehps$ is $\alltasks{\somehps} = \bigcup_{i \in \set{1,\ldots,n}} \tasks_i$, and the \emph{set of shared activities} of $\somehps$ is  $\ctasks{\somehps} = \bigcap_{i \in \set{1,\ldots,n}} \tasks_i$. $\somehps$ accepts trace $\trace \in (\alltasks{\somehps})^*$ if, for every $i \in \set{1,\ldots,n}$, we have that specification $\tup{\tasks_i,\proc_i}$ accepts $\proj{\trace}{\tasks_i}$ (in the sense of Definition~\ref{def:ps}).
\end{definition}

\cref{def:hps} leads to relatively simple semantics where a trace is accepted by the process frame iff each specification therein accepts the projection of that trace on the alphabet of that specification, which is equivalent to synchronous execution of each activity over all specifications containing that activity. This has the benefit of enabling arbitrary combinations of (non-conflicting) process specifications to be executed, but comes with significant implications on automated process (frame) discovery. These implications are further discussed in \cref{sec:discovery_implications}.

Note that \cref{def:hps} can be grounded using any languages that allow for reasoning in terms of accepted and non-accepted activity sequences, including Petri nets~\cite{DBLP:journals/pieee/Murata89,DBLP:journals/topnoc/HeeSW13a} and \declare~\cite{DBLP:conf/edoc/PesicSA07}, which are used in all following sections as representative examples of procedural and declarative modeling paradigms, respectively. In addition, the definitions of this section allow the same process frame to be represented using different sets of specifications. For example, the set of traces accepted by variants of the same process frame, obtained by gluing together or splitting the specifications therein, will remain identical. This allows for different levels of compositionality and, crucially for this paper, enables choosing between different (procedural and declarative) modeling languages on a per specification basis.

\subsection{Minimalistic Example}
\label{sec:example}

\begin{figure}[t]
    \centering
    \includegraphics[width=.94\textwidth]{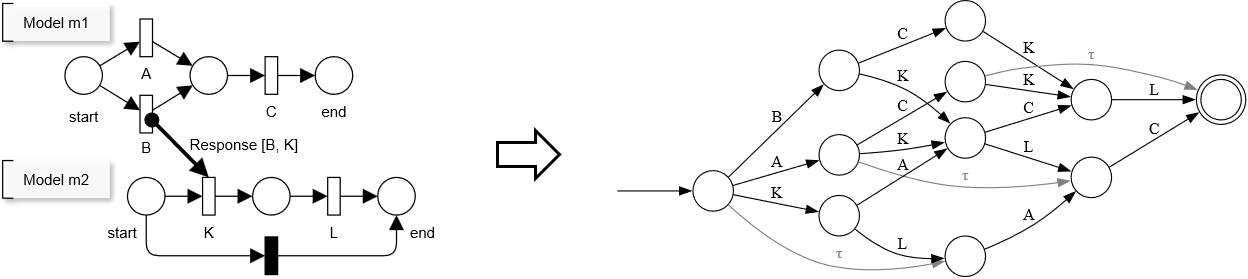}
    \caption{Minimalistic example of a multi-model ABPMS process frame and the resulting state-space (taken from~\cite{Maggi2026}).}
    \label{fig:interplay}
\end{figure}

In its simplest form, our conceptualization of the process frame is a parallel composition of individual specifications, where each specification may use a different modeling language and may follow a different modeling paradigm. However, shared activities (and compositionality) of the multi-model process frame can be leveraged to express various interdependencies between individual specifications. As an example, consider the process frame given in \cref{fig:interplay} (left). It consists of Petri nets $m1$ and $m2$, and a \declare constraint \constraint{Response}(\activity{B}, \activity{K}), with the latter specifying that each execution of \activity{B} (in $m1$) must be followed by at least one \activity{K} (in $m2$). This declarative constraint, along with the procedural specifications $m1$ and $m2$, leads to an overall behavior where, contrary to usual parallel constructs, activities of different specifications cannot be arbitrarily mixed. That is, the possible \emph{interleavings} of the given specifications are \emph{constrained} by the combination of execution requirements imposed on the shared activities by each individual specification, in this case, affecting how $m2$ is executed in relation to $m1$, but also limiting how \constraint{Response}(\activity{B}, \activity{K}) can be satisfied~\cite{ALMAN2023102512}.

More specifically, $m2$ requires only that, if \activity{K} is executed, then \activity{L} must be executed next (while not precluding any activities of $m1$ in between). This is similar to the \constraint{Response} constraint, but with an additional \emph{cardinality} requirement of 0...1 on both \activity{K} and \activity{L}, such that both activities must be executed an equal number of times. Furthermore, if \activity{B} is executed (in $m1$), then \activity{K} (in $m2$) becomes required due to the \constraint{Response} constraint between these activities, indirectly causing \activity{L} (in $m2$) to also become required due to the structure of $m2$. Conversely, note that the \constraint{Response} constraint itself does not impose any upper bound on the number of times \activity{B} and \activity{K} could be executed, while the given Petri nets do not allow these activities to be executed more than once. As a result, if the overall execution starts with \activity{K}, then \activity{B} cannot be executed (in $m1$) as this would cause the \constraint{Response} constraint to require repeating \activity{K}, which is not allowed by $m2$. The corresponding complete behavior is given, as a deterministic finite state automaton, in \cref{fig:interplay} (right).

\subsection{Implications on Discovery}
\label{sec:discovery_implications}

The multi-model process frame, as introduced in this paper, does not necessarily exclude the use of existing process discovery approaches. Such approaches need to produce models that are unambiguous, consistent and executable, while also avoiding conflicts between the models that are used as specifications in the same process frame. Ensuring the latter can already present a notable challenge, but one should also consider the conceptual differences between process models and process frames (cf. \cref{sec:conceptualization}). In the following, we describe four main implications arising from these differences, focusing on the most relevant ones for the discovery of a multi-model process frame.

First, all process discovery approaches rely on historical execution data, which may contain noise, infrequent behaviors, systematic mistakes, etc. As a result, automated process discovery alone is unlikely to yield a complete and accurate process frame. Instead, it should be positioned as an approach for discovering an initial process frame, to be adjusted according to the expected behavioral boundaries of the ABPMS. One can take multiple strategies here. For example, first discovering an overview of the underlying processes, and then manually adding details, such as behaviors that are uncommon, but nevertheless valid. Another option, following the open-world assumption, is to first discover a relatively permissive representation, which allows all observed behavior, and then manually constraining the behaviors outside the expected behavioral boundaries of a given ABPMS. The process frame, as introduced in this paper, is especially suitable for the latter, as additional specifications can always be added to constrain the behaviors that would be otherwise allowed by the specifications already present in the process frame (cf. \cref{sec:conceptualization}).

Second, the ABPMS manifesto explicitly highlights that an ABPMS may manage more than a single process~\cite{DBLP:journals/tmis/DumasFLMMRACGFGRVW23}. However, the majority of existing process discovery approaches are developed with a focus on discovering a single isolated process. This includes contemporary approaches, such as Heuristics Miner~\cite{DBLP:journals/icae/WeijtersA03}, Inductive Miner~\cite{DBLP:conf/bpm/LeemansFA13}, and Split Miner~\cite{DBLP:journals/kais/AugustoCDRP19}, as well as, hybrid approaches, such as Fusion Miner~\cite{DBLP:journals/dss/SmedtWV15} and the discovery of hierarchical models in~\cite{DBLP:conf/bpm/MaggiSR14}. A similar argument can also be made for object-centric approaches, e.g., \cite{DBLP:conf/sefm/Aalst19,DBLP:journals/fuin/AalstB20,DBLP:conf/icpm/DettenSL24,DBLP:conf/bpm/KustersA25}, where the notion of a business process is blurred by instead focusing on activity executions over the relevant object types and instances, with no corresponding hybrid discovery approaches yet developed. While one could create representative objects for individual processes, this would also require explicit foreknowledge of the exact activity-process relations, which is not assumed in this paper. Furthermore, recall that each specification of the multi-model process frame is intended to represent some logical component of the overall behavioral boundary of an ABPMS (cf. \cref{sec:conceptualization}), and is, therefore, not necessarily tied to specific processes nor objects therein.

\begin{figure}[t]
    \centering
    \includegraphics[width=.28\textwidth]{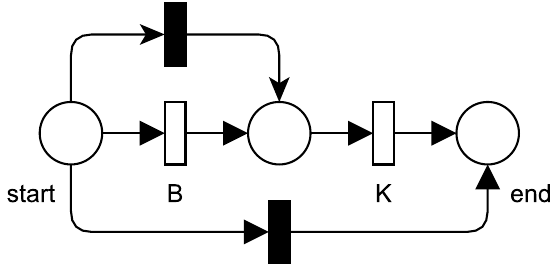}
    \caption{Explicit representation of the the relation between \activity{B} and \activity{K} in \cref{fig:interplay}.}
    \label{fig:relation_B_K}
\end{figure}

Third, the compositionality of specifications within the process frame, as introduced in this paper, can become relatively complex to understand. Recall, for example, that starting the execution of the process frame given in \cref{fig:interplay} with activity \activity{K} will preclude executing \activity{A} in the same activity sequence. This is similar to the well-known problem of hidden dependencies in the \declare language~\cite{DBLP:journals/is/SmedtWSV18,DBLP:conf/caise/HaisjacklZSHRPW13}, where some behavior becomes implicitly forbidden due to how individual constraints in the overall model interact. The aforementioned relation between \activity{B} and \activity{K} can be made explicit by adding the additional specification given in \cref{fig:relation_B_K} to the process frame given in \cref{fig:interplay}. As highlighted in \cref{sec:formalization}, this addition would not affect the overall behavior of the ABPMS, but it would nevertheless be beneficial to discover from the representational point of view and, notably, an event log of the process frame in \cref{fig:interplay} is likely to contain all the necessary information for doing so.

Fourth, and most importantly for this paper, the semi-concurrency of the process frame, as outlined in this paper, complicates the use of existing (procedural) process discovery approaches, including the approaches already mentioned in this section, because these approaches rely on detecting frequent directly-follows relations between pairs of activities. Consider again the process frame in \cref{fig:interplay}. From the perspective of $m2$, every execution of \activity{K} must be followed by \activity{L}. However, if \activity{C} (in $m1$) is executed between \activity{K} and \activity{L}, then that will ``break'' the corresponding directly-follows relation in the overall execution trace, while leaving the projection of that trace on the activities of $m2$ intact (cf. \cref{sec:formalization}). If all behavioral components are known in advance, then this issue can be solved by simply splitting the event log into corresponding sub-logs, similar to the object-centric approach in \cite{DBLP:journals/fuin/AalstB20}, however, such foreknowledge cannot be assumed in all cases. Furthermore, splitting the event log according to a fixed set of components from the outset would not support the exploration of alternative process frame variants, that may be obtained, for example, by replacing some subsets of declarative constraints with behaviorally equivalent procedural fragments (or vice versa) or, as already discussed, by adding additional specifications to reveal otherwise implicit interactions between specifications.

\section{Procedural Behaviors Under the Open-World Assumption}
\label{sec:behaviors}

The multi-model process frame, as introduced in \cref{sec:frame}, has multiple similarities to the design of the declarative process modeling language \declare~\cite{DBLP:conf/edoc/PesicSA07}. Both consist of components\footnote{Specifications, in case of the multi-model process frame, and constraints, in case of \declare.} that: follow the open-world assumption; can be combined based on shared activities; and are interpreted semi-concurrently over global execution traces. These characteristics are inherently supported by existing \declare process discovery approaches (e.g.,~\cite{DBLP:journals/tmis/CiccioM15,DBLP:journals/is/MaggiCFK18}), but have not been relevant for procedural approaches (e.g.,~\cite{DBLP:journals/kais/AugustoCDRP19,DBLP:conf/bpm/LeemansFA13,DBLP:journals/icae/WeijtersA03}). This, along with most \declare constraints representing various eventually-follows, rather than directly-follows, relations (cf. \cref{sec:discovery_implications}), makes \declare-based process discovery a good starting point for the automated discovery of a multi-model process frame.

The general idea would be to first discover a set of \declare constraints that hold over most, or even all, observed behaviors in a given event log, thus giving a fully declarative variant of the process frame. As discussed in \cref{sec:formalization}, any components of this process frame, i.e., subsets of discovered constraints, can then be replaced with procedural specifications, assuming that such specifications are behaviorally equivalent under the open-world assumption, i.e., accept exactly the same set of trace projections that the replaced constraints accept. In some sense, this reverses the well-known ``pockets of flexibility'' approach~\cite{DBLP:conf/er/SadiqSO01} into ``pockets of rigidity'', changing a fully declarative model into a hybrid business process representation by using procedural constructs for the more rigid sub-behaviors within it.

The above assumes that corresponding procedural specifications can be derived from (discovered) \declare constraints. We have previously demonstrated that this is technically feasible when one assumes no activity repetitions~\cite{DBLP:conf/pmai/AlmanCGMM24}. In this section, we demonstrate that this is, in most cases, also feasible for repeatable procedural behaviors, assuming some extensions of the \declare language. To do so, we leverage deterministic finite state automata (DFAs), as described in~\cite{DBLP:conf/bpm/Westergaard11} specifically, to represent the state-space of various (sets of) \declare constraints. On the one hand, the corresponding implementation, called LTL2Automaton\footnote{Available as part of the process mining platform ProM (\url{https://promtools.org/}).}, provides convenient methods for constructing and combining DFAs of any \declare constraints. On the other hand, this specific implementation supports ``negations'' and ``any transitions'', with the former matching all input symbols (in our case, activity names) not listed in the negation, and the latter, as the name suggests, matching any input symbol. This allows constructing DFAs with a defined behavior for all symbols, including the ones that are not explicitly used in that DFA, which, in turn, allows for DFAs that handle the trace projection, as defined in \cref{sec:formalization}, through appropriate self-loops on the DFA states.

In the following, \cref{sec:declare_analysis} presents an analysis and our extensions of the \declare language for detecting repeatable behaviors, \cref{sec:isolated_behaviors} demonstrates the manifestation of common procedural behaviors as sets of \declare constraints, and \cref{sec:overlapping_behaviors} demonstrates that more complex procedural structures can also be accurately detected.

\subsection{Analysis and Extensions of the Declare Language}
\label{sec:declare_analysis}

As part of our earlier work~\cite{DBLP:conf/pmai/AlmanCGMM24}, we presented a mapping of \declare constraints into fragments of Petri nets and an initial approach for creating larger Petri nets out of these fragments. Both of these contributions assume that activity executions and choices are never repeated. If that assumption does not hold, then the aforementioned mapping and approach are no longer reliable. To recap, the following \declare constraints were used\footnote{As usual, \activity{A} and \activity{B} serve as activity placeholders when describing \declare constraints in general terms.}:
\begin{itemize}
    \setlength\itemsep{0em}
    \item \constraint{Response}(\activity{A}, \activity{B}) -- If \activity{A} occurs, then it must be eventually followed by \activity{B};
    \item \constraint{Precedence}(\activity{A}, \activity{B}) -- \activity{B} can occur only if \activity{A} has previously occurred;
    \item \constraint{Succession}(\activity{A}, \activity{B}) -- If \activity{A} occurs, then it must be eventually followed by \activity{B} and \activity{B} cannot occur before the first occurrence of \activity{A} (this is equivalent to combining \constraint{Response} and \constraint{Precedence} on the same activities);
    \item \constraint{Not Co-Existence}(\activity{A}, \activity{B}) -- If \activity{A} occurs, then \activity{B} must not occur;
    \item \constraint{Co-Existence}(\activity{A}, \activity{B}) -- If \activity{A} occurs, then \activity{B} must also occur;
    \item \constraint{Not Succession}(\activity{A}, \activity{B}) -- If \activity{A} occurs, then no \activity{B} may occur afterward.
\end{itemize}

All constraints of the \declare language are interpreted globally~\cite{DBLP:conf/bpm/AlmanCMMA21}, meaning that a constraint must hold over the entire sequence of executed activities, i.e., over the entire (global) trace. For example, \constraint{Not Co-Existence}(\activity{A}, \activity{B}), which may seem like a natural fit for detecting exclusive choices, means that the entire trace can only contain \activity{A} or \activity{B}, or neither, but never both. If there are repeatable behaviors, for example, a loop that allows an exclusive choice to be made multiple times, then a \constraint{Not Co-Existence} constraint would hold (over the entire trace) only if there is a dependency between the loop iterations, such that the activity, which was chosen in the first loop iteration, must also be chosen in all the following iterations. In general, declarative handling of repetitions, at least in case of \declare, is not analogous to the handling of repetitions in common procedural process modeling languages, motivating the analysis and the extensions presented in the following sub-sections.

\subsubsection{Repeatable Behaviors in \declare}
\label{sec:declare_loops}

\declare provides several mechanisms for handling activity repetitions. For example, most unary constraints specify how many times a specific activity can (e.g., \constraint{Absence3}) or must (e.g., \constraint{Existence2}) occur. In addition, there are constraints for the exact number of activity occurrences (e.g., \constraint{Exactly2}). However, as with all \declare constraints, the resulting \emph{activity cardinalities} would apply over the entire execution trace~\cite{DBLP:conf/bpm/AlmanCMMA21}, which is not always desirable in the presence of repetitions. For example, it is not possible to specify that \activity{A} must occur at least twice \emph{per loop iteration}, as the constraint \constraint{Existence2}(\activity{A}) would apply over all iterations. Consequently, unary constraints are not reliable for detecting procedural behaviors (although they can still be included in the resulting process frame, if discovered). Beyond unary constraints, the only other \declare constraints, that can handle activity repetitions natively, are:
\begin{itemize}
    \setlength\itemsep{0em}
    \item \constraint{Alternate Succession}(\activity{A}, \activity{B}) -- \activity{A} and \activity{B} occur in the same trace if and only if every occurrence of \activity{A} is eventually followed by an occurrence of \activity{B}, and \activity{A} and \activity{B} strictly alternate, i.e., each occurrence of \activity{A} is eventually followed by exactly one \activity{B} before either the trace ends or \activity{A} reoccurs;
    
    \item \constraint{Alternate Response}(\activity{A}, \activity{B}) -- each occurrence of \activity{A} must be eventually followed by at least one occurrence of \activity{B} and there must be at least one occurrence of \activity{B} between any pair of \activity{A} occurrences;
    
    \item \constraint{Alternate Precedence}(\activity{A}, \activity{B}) -- each occurrence of \activity{B} must be preceded by at least one occurrence of \activity{A} and there must be at least one occurrence of \activity{A} between any pair of \activity{B} occurrences.
    
\end{itemize}

\begin{figure*}[t]
	\centering
	\subfigure[\constraint{Not Co-Existence}(\activity{A},\activity{B}).]{\includegraphics[width=0.3\textwidth]{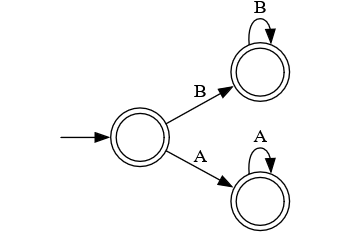} \label{fig:aut_not_coex}}\hfil
	\subfigure[\constraint{Co-Existence}(\activity{A},\activity{B}).]{\includegraphics[width=0.3\textwidth]{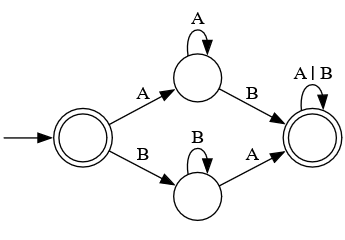}
		\label{fig:aut_coex}}\hfil
	\subfigure[\constraint{Not Succession}(\activity{A},\activity{B}).]{\includegraphics[width=0.3\textwidth]{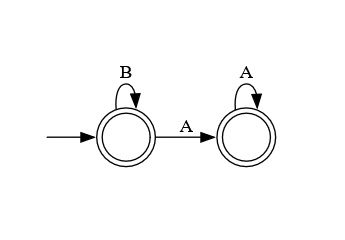}
		\label{fig:aut_not_succ}}
	\caption{DFA representations of the \declare templates used in~\cite{DBLP:conf/pmai/AlmanCGMM24} that do not have corresponding alternate-type templates. Arcs leading to non-accepting trap states are omitted for brevity. Furthermore, activities not listed on any of the DFA arcs are processed by staying in the current state (i.e., as implicit self-loops).
}
	\label{fig:automata_noAlt}
\end{figure*}

This provides alternatives to the \constraint{Succession}, \constraint{Response}, and \constraint{Precedence} constraints used in~\cite{DBLP:conf/pmai/AlmanCGMM24}. However, the \declare language does not have ``alternate versions'' for \constraint{Not Co-Existence}, \constraint{Co-Existence}, and \constraint{Not Succession}, and none of these can be reliably used for detecting procedural behaviors within the context of repeatable behaviors. This is apparent from the corresponding DFAs (cf. \cref{fig:automata_noAlt}), when, for example, assuming that \activity{A} and \activity{B} are activities within some outer-loop. For example, \constraint{Co-Existence}(\activity{A}, \activity{B}), which may seem like a natural fit for detecting parallelism, would also be discovered if \activity{A} is executed in the first and \activity{B} in the second loop iteration, while \constraint{Not Succession}(\activity{A}, \activity{B}) would only be discovered if there is a dependency between the outer-loop iterations, such that an iteration containing \activity{A} can never be followed by an iteration containing \activity{B}.

There are a few possibilities for circumventing these issues. One option involves identifying all repeating sub-sequences of activities (e.g., by checking for repeated activity executions), extracting each repetition as a separate trace, and discovering \declare constraints from these extracted traces. Another option is to augment \declare constraints with a notion of scope, similar to~\cite{DBLP:conf/icse/DwyerAC99}. However, we aim to keep the modifications of the \declare language to a minimum, which we achieve by using \constraint{Not Chain Succession} constraints. \constraint{Not Chain Succession} is already part of the \declare language and allows identifying, for each activity, the activities that can and cannot be executed immediately after that activity (substituting for \constraint{Co-Existence} and \constraint{Not Co-Existence}, respectively), without causing dependencies between loop iterations. The exact use of \constraint{Not Chain Succession} is discussed further in \cref{sec:cardinalities,sec:isolated_behaviors}.

\subsubsection{Additional Constraint Templates}
\label{sec:constraints}

\cref{tab:constraint_cardinalities} gives an overview of all \declare constraints used in this paper for detecting common procedural behaviors. In addition to the ones already discussed, we use two conjunctions of constraints, one ternary constraint (i.e., a constraint involving three activities) from literature, and one novel ternary constraint.

\begin{figure*}[t]
	\centering
	\subfigure[\constraint{Succession}(\activity{A},\activity{B}).]{\includegraphics[width=0.32\textwidth]{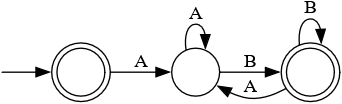} \label{fig:aut_succ}}\hfil
	\subfigure[\constraint{Alternate Response}(\activity{A},\activity{B}).]{\includegraphics[width=0.32\textwidth]{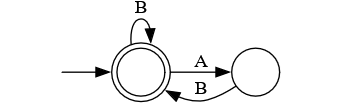}
		\label{fig:aut_altResp}}\hfil
	\subfigure[\constraint{Succession}(\activity{A},\activity{B}) + \newline \constraint{Alternate Response}(\activity{A},\activity{B}).]  {\includegraphics[width=0.32\textwidth]{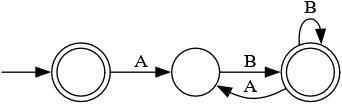}
		\label{fig:aut_succ_altResp}}\hfil
	\subfigure[\constraint{Interposition}(\activity{A},\activity{B},\activity{C}).]  {\includegraphics[width=0.42\textwidth]{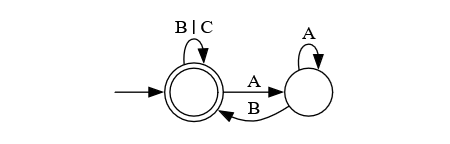}
		\label{fig:aut_br_enf}}\hfil
	\subfigure[\constraint{Balanced Enablement}(\activity{A},\activity{B},\activity{C}).]  {\includegraphics[width=0.42\textwidth]{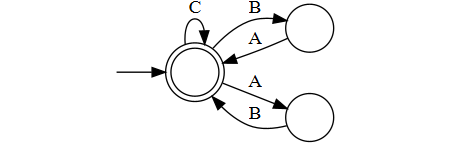}
		\label{fig:aut_bl_enable}}
	\caption{DFAs of the relation \constraint{Alternate Response}(\activity{A}, \activity{B}) $+$ \constraint{Succession}(\activity{A}, \activity{B}) and the ternary templates \constraint{Interposition}(\activity{A}, \activity{B}, \activity{C}) and \constraint{Balanced Enablement}(\activity{A}, \activity{B}, \activity{C}).
}
	\label{fig:automata_card_ex}
\end{figure*}

The conjunctions are used for detecting relations between activities that cannot be captured by a single \declare constraint alone. In particular, the conjunction \constraint{Alternate Response}(\activity{A}, \activity{B}) + \constraint{Succession}(\activity{A}, \activity{B}) is used to detect the start of repeatable regions (i.e., loops), where the repeatable region must be executed at least once. Conversely, \constraint{Alternate Precedence}(\activity{A}, \activity{B}) + \constraint{Succession}(\activity{A}, \activity{B}) is used to detect the end of such regions. The behavior of either conjunction is equivalent to the cross-product of the corresponding DFAs. As an example, \cref{fig:aut_succ,fig:aut_altResp,fig:aut_succ_altResp} show the corresponding individual DFAs, and their cross-product, for the conjunction \constraint{Alternate Response}(\activity{A}, \activity{B}) + \constraint{Succession}(\activity{A}, \activity{B}).

We include the ternary constraint \constraint{Interposition}(\activity{A}, \activity{B}, \activity{C}) from~\cite{DBLP:series/lnbip/Montali10}. This constraint specifies that whenever \activity{A} occurs, \activity{C} cannot occur until \activity{B} has occurred. For example, the sequence \activity{A}, \activity{B}, \activity{C} is allowed, while the sequence \activity{A}, \activity{C} is not. This constraint does not require neither \activity{A} nor \activity{C} to occur, but if \activity{A} does occur, then \activity{B} becomes required. It can also be interpreted as follows: \activity{C} is initially enabled (but not required); the occurrence of \activity{A} disables \activity{C} and requires \activity{B}; and \activity{C} becomes re-enabled (but not required) after \activity{B} occurs. The corresponding DFA is given in \cref{fig:aut_br_enf}. The \constraint{Interposition} constraint is used in \cref{sec:isolated_behaviors} to detect certain variants of OR choices.

We also introduce a novel ternary constraint, which we call \constraint{Balanced Enablement}(\activity{A}, \activity{B}, \activity{C}). It specifies that: at any point during execution, the absolute difference between the frequencies of \activity{A} and \activity{B} must be at most one; \activity{C} may only occur when the frequencies of \activity{A} and \activity{B} are equal; and the frequencies of \activity{A} and \activity{B} must be equal at the end of the trace. The \constraint{Balanced Enablement} constraint is used in \cref{sec:isolated_behaviors} to detect three variants of parallelism.

While technically not an extension of the \declare language, it is important to note that Declare Miner~\cite{DBLP:journals/is/MaggiCFK18}, which is used in \cref{sec:isolated_behaviors,sec:overlapping_behaviors}, does not discover binary constraints where both activities are the same. This is reasonable for most \declare constraints, e.g., \constraint{Succession}(\activity{A}, \activity{A}) implies an infinite trace, while \constraint{Co-Existence}(\activity{A}, \activity{A}) is always satisfied and, therefore, not meaningful. The exception, in the context of this paper, is \constraint{Not Chain Succession}. Specifically, \constraint{Not Chain Succession}(\activity{A}, \activity{A}) indicates that \activity{A} cannot be immediately followed by itself, i.e., there cannot be a self-loop on \activity{A}, but \activity{A} may still be repeated later. Such constraints can be easily discovered by checking the given event log for any consecutive occurrences of the same activity.

\subsubsection{Relative Activity Cardinalities}
\label{sec:cardinalities}

\begin{table}[t]
\scriptsize
\centering
\begin{tabular}{l|l|l}
\textbf{Constraint Template}                                                                  & \textbf{Relative Cardinalities}                                                                 & \textbf{Related Procedural Construct}                                                 \\ \hline
Alternate Succession{(}A,B{)}                                                                 & \begin{tabular}[c]{@{}l@{}}Outgoing 1...1 cardinality\\ Incoming 1...1 cardinality\end{tabular} & \begin{tabular}[c]{@{}l@{}}Mandatory sequence flow\\ Mandatory AND\end{tabular}        \\ \hline
\begin{tabular}[c]{@{}l@{}}Succession{(}A,B{)} +\\ Alternate Response{(}A,B{)}\end{tabular}   & Outgoing 1...n cardinality                                                                      & \begin{tabular}[c]{@{}l@{}}Start of loop\\ (one iteration required)\end{tabular}        \\ \hline
\begin{tabular}[c]{@{}l@{}}Succession{(}A,B{)} +\\ Alternate Precedence{(}A,B{)}\end{tabular} & Incoming 1...n cardinality                                                                      & \begin{tabular}[c]{@{}l@{}}End of loop\\ (one iteration required)\end{tabular} \\ \hline
Alternate Precedence{(}A,B{)}                                                                 & \begin{tabular}[c]{@{}l@{}}Outgoing 0...1 cardinality\\ Incoming 1...n cardinality\end{tabular}   & \begin{tabular}[c]{@{}l@{}}Start of XOR\\ Start of optional process flow\end{tabular} \\ \hline
Alternate Response{(}A,B{)}                                                                   & \begin{tabular}[c]{@{}l@{}}Outgoing 1...n cardinality\\ Incoming 0...1 cardinality\end{tabular}   & \begin{tabular}[c]{@{}l@{}}End of XOR\\ End of optional process flow\end{tabular}     \\ \hline
Precedence{(}A,B{)}                                                                           & Outgoing 0...n cardinality   & \begin{tabular}[c]{@{}l@{}}Start of loop\\ (no iterations required)\end{tabular}        \\ \hline
Response{(}A,B{)}                                                                             & Incoming 0...n cardinality   & \begin{tabular}[c]{@{}l@{}}End of loop\\ (no iterations required)\end{tabular}         \\ \hline
Succession{(}A,B{)}                                                                           & \textit{Not used for cardinality}   & \begin{tabular}[c]{@{}l@{}}Independently repeatable activities\\ (one occurrence of both required) \end{tabular}        \\ \hline
Not Chain Succession{(}A,B{)}                                                                             & \textit{Not used for cardinality}   & \begin{tabular}[c]{@{}l@{}}All XOR and inclusive OR types\\ \textit{Supports operationalizing all patterns}\end{tabular}         \\ \hline
Interposition{(}A,B,C{)}                                                                             & \textit{Not used for cardinality}   & \begin{tabular}[c]{@{}l@{}}Inclusive OR (mandatory and optional)\end{tabular}         \\ \hline
Balanced Enablement{(}A,B,C{)}                                                                             & \textit{Not used for cardinality}   & \begin{tabular}[c]{@{}l@{}}All AND types, except mandatory\end{tabular}
\end{tabular}
\caption{Cardinalities of \declare templates and their relations to procedural constructs.}\label{tab:constraint_cardinalities}
\end{table}

Binary \declare constraints are, in general, interpreted as expressing some form of co-existence and/or temporal ordering between activity pairs. However, as highlighted in~\cite{DBLP:journals/is/SmedtWSV18}, they also impose bounds on the \emph{relative frequencies} between activity pairs. For example, \constraint{Alternate Succession}(\activity{A}, \activity{B}) imposes that every \activity{A} must be followed by exactly one \activity{B}, and \activity{A} cannot recur before the required \activity{B} has occurred, leading to a 1...1 cardinality, where each \activity{A} corresponds to exactly one \activity{B}, and vice versa. We refer to such relations as \emph{relative activity cardinalities} and, as a further refinement, we distinguish \emph{outgoing} and \emph{incoming cardinalities} based on the temporal ordering imposed by the constraint. Specifically, the outgoing cardinality refers to the number of times the follower activity can occur after the predecessor activity has occurred, while the incoming cardinality refers to how many times the predecessor can occur before the follower occurs.

For example, \constraint{Alternate Response}(\activity{A}, \activity{B}) (cf.~\cref{fig:aut_altResp}) imposes an outgoing cardinality of 1...n, meaning that each occurrence of \activity{A} is followed by one or more occurrences of \activity{B}, and an incoming cardinality of 0...1, meaning that each occurrence of \activity{B} is preceded by zero or one occurrences of \activity{A}. The relative activity cardinalities used in this paper are summarized in \cref{tab:constraint_cardinalities}. Note that some cardinalities can be more complex. For instance, the incoming cardinality of \constraint{Succession}(\activity{A}, \activity{B}) (cf.~\cref{fig:aut_succ}) is 1...n for the first occurrence of \activity{B}, but 0...n for all subsequent occurrences of \activity{B}.

Combining relative activity cardinalities with the temporal orderings of constraints allows identifying \emph{predecessor} and \emph{follower} sets of activities, between which, the execution cardinality of the process changes. A limitation here is that every activity in a predecessor group (e.g., activities before a loop) will have the same relative cardinality with all activities in the follower group (e.g., activities inside the loop body). The precise point where this cardinality changes (e.g., entry point of the loop body) can be pinpointed through \constraint{Not Chain Succession} constraints between the groups. In particular, \constraint{Not Chain Succession} constraints will exist from all activities in the predecessor group to all activities in the follower group, except for those directly preceding and succeeding the point at which the cardinality changes.

\subsection{Common Procedural Behaviors}
\label{sec:isolated_behaviors}

\begin{table}[t]
\scriptsize
\centering
\begin{tabular}{r|cccc}
                     & Sequence & AND & XOR & OR \\ \hline
Mandatory~~~1...1            & +             & +                 & +                      & +*                    \\
Optional~~~0...1             & +             & +*                & +                      & +*                    \\
Mandatory Repeatable~~~1...n & +             & +*                & +                      & -                     \\
Optional Repeatable~~~0...n  & +             & +*                & +                      & -                    
\end{tabular}
\caption{Discoverability of procedural behaviors through \declare constraints. Asterisk denotes behaviors that require the extensions described in \cref{sec:constraints}.}\label{tab:manifestation}
\end{table}

In this section, we demonstrate the feasibility of detecting common procedural behaviors from sets of discovered \declare constraints. We focus on the same procedural behaviors as in~\cite{DBLP:conf/pmai/AlmanCGMM24}, namely, sequence flow; parallelism (AND); exclusive choice (XOR); and inclusive choice (OR). Each behavior, as a whole, is considered with respect to four cardinalities: 1...1, referred to as \emph{mandatory}; 0...1, referred to as \emph{optional}; 1...n, referred to as \emph{mandatory repeatable}; and 0...n, referred to as \emph{optional repeatable}. This results in 16 procedural variants (cf. \cref{tab:manifestation}), which we consider in the context of a larger repeatable process by placing the procedural structures within larger outer-loops (e.g., \cref{fig:pn_seq}).

We take an approach inspired by the process rediscovery problem~\cite{DBLP:books/daglib/0027363}. More specifically, we first create a Petri net of the procedural behavior being considered, then we generate an artificial event log based on the model, discover a \declare model from that event log, validate that the DFA representation of the discovered \declare model corresponds to the procedural behavior being considered, and, if so, determine the relevant \declare constraints that have to be discovered for this procedural behavior to be detected correctly. The artificial event logs are created, using~\cite{DBLP:conf/bpm/AlmanMMR23}, such that they include at least one example of each distinct activity sequence that is allowed and no activity sequences that are not allowed by the given Petri net, with an upper bound on the number of loop iterations.

The \declare models are discovered using a specifically developed prototype\footnote{Publicly available at: \url{https://github.com/antialman/multi-model-miner}}, that uses the Declare Miner~\cite{DBLP:journals/is/MaggiCFK18} internally, but with adjustments for the additional constraints considered in this paper (cf. \cref{sec:constraints}). All constraints are discovered using 100\% support threshold, but without applying vacuity detection~\cite{DBLP:journals/sttt/KupfermanV03}. This means that a constraint is discovered between a pair of activities if there is at least one positive example and no counterexamples of that constraint. Pruning of redundant constraints~\cite{DBLP:conf/caise/MaggiBA13} is not applied, as it is not available for \constraint{Alternate Succession}, \constraint{Alternate Response}, and \constraint{Alternate Precedence} constraints, nor for the ternary constraints considered in this paper.

In the following, we describe the results for each procedural behavior, including the four cardinalities. For sequence flows, we give the Petri nets of all cardinalities explicitly (cf. \cref{fig:pn_seq}), while for the other procedural behaviors, we give only the figure for \emph{optional repeatable} cardinality, as the other cardinalities can be easily obtained from it. In addition, we show the DFA representation of the discovered \declare constraints for all 16 variants and describe the constraints that must (or must not) hold for the corresponding mappings.\footnote{All given DFAs accept an empty trace, which can be addressed by simply checking if the given event log contains any empty traces.} As shown in \cref{tab:manifestation}, standard \declare constraints allow detecting 9 out of the 16 variants, which rises to 14 out of 16, when including the additions from \cref{sec:constraints}. The remaining two variants, mandatory repeatable and optional repeatable OR, would require additional post-processing.

\subsubsection{Sequence Flow}
\label{sec:seq_flow}

\begin{figure*}[t]
	\centering
	\subfigure[Mandatory sequence flow.]{\includegraphics[width=0.48\textwidth]{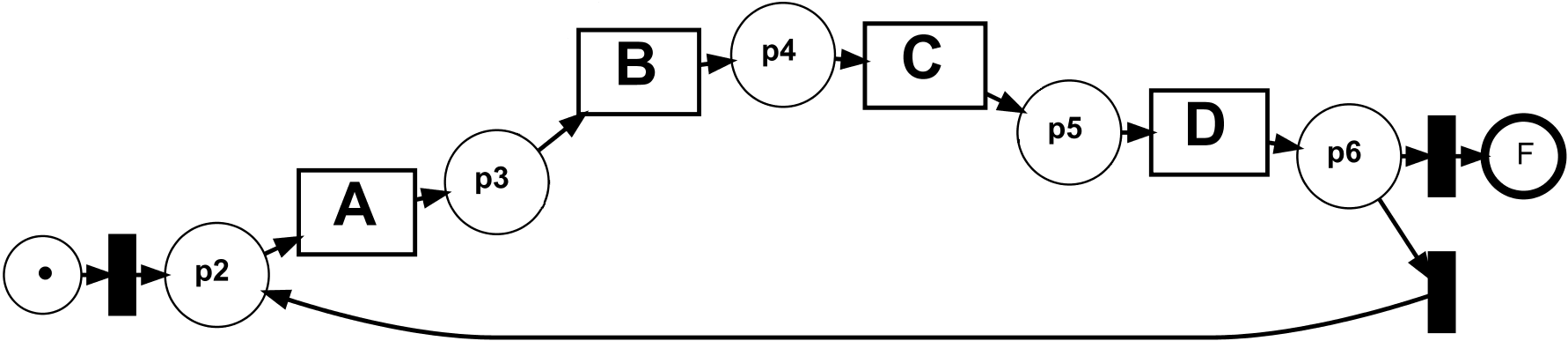} \label{fig:pn_seq_simple}}\hfil
	\subfigure[Optional sequence flow.]{\includegraphics[width=0.48\textwidth]{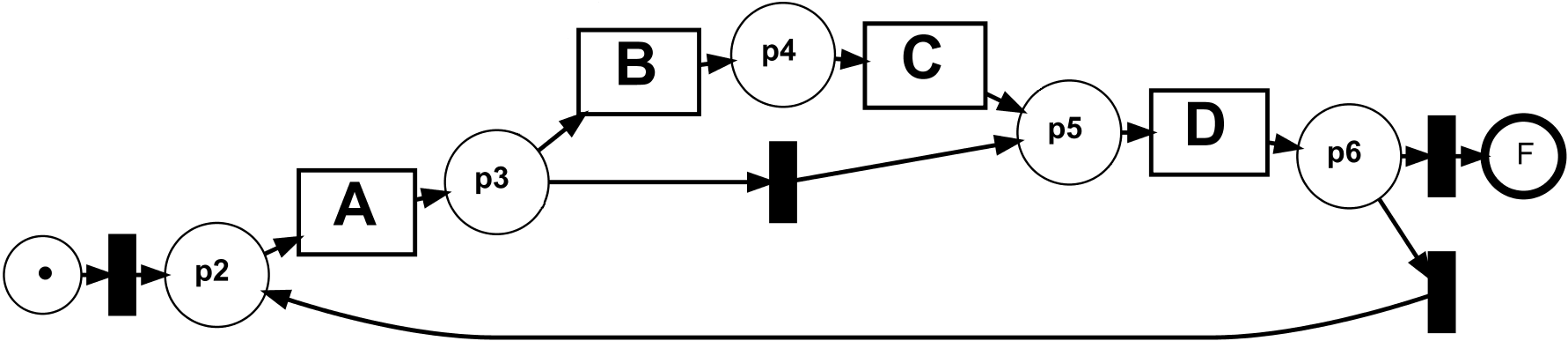}
		\label{fig:pn_seq_optional}}\hfil
	\subfigure[Mandatory repeatable sequence flow.]  {\includegraphics[width=0.48\textwidth]{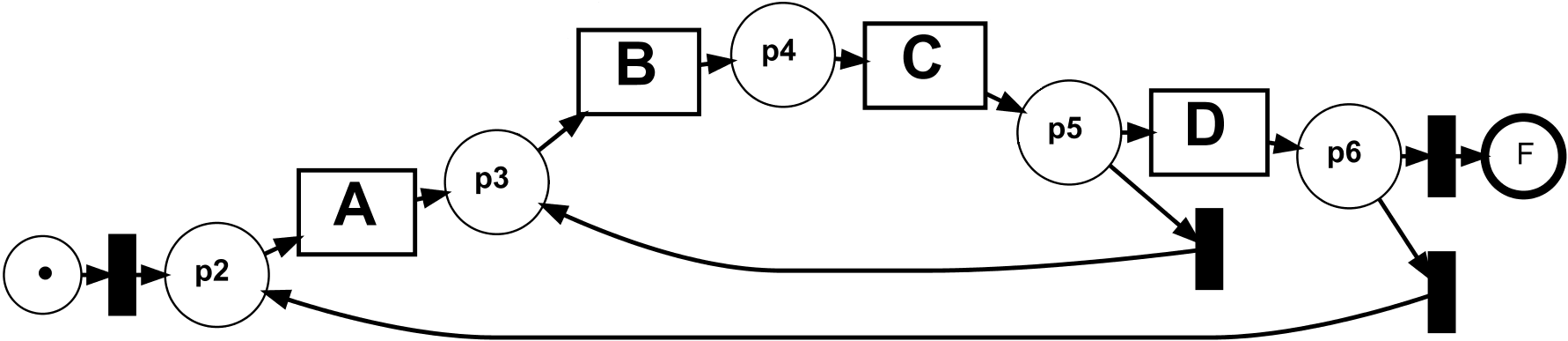}
		\label{fig:pn_seq_repeatable}}\hfil
	\subfigure[Optional repeatable sequence flow.]  {\includegraphics[width=0.48\textwidth]{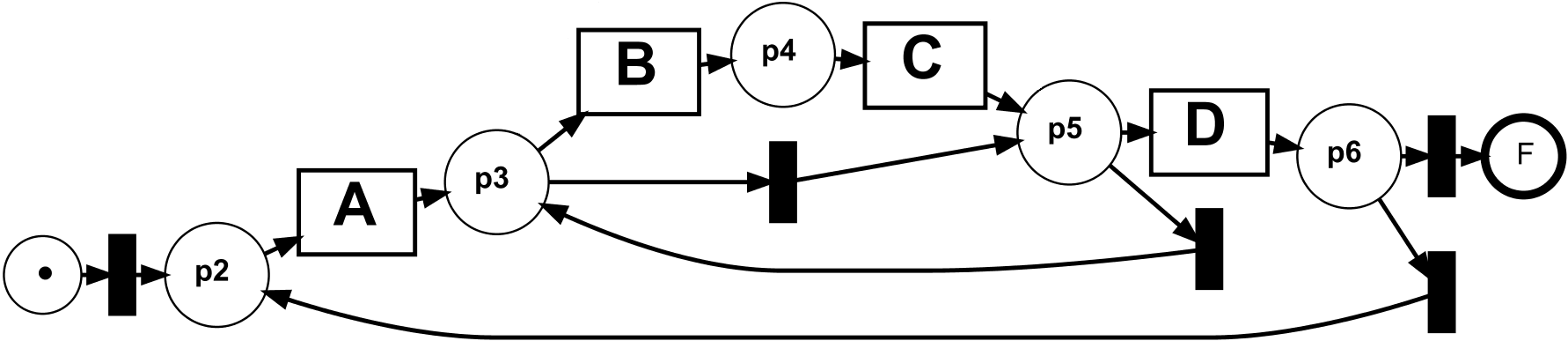}
		\label{fig:pn_seq_optional_repeatable}}
	\caption{Petri nets illustrating the four variants of the sequence flow pattern. The layout highlights the use of silent transitions to represent optionality and repeatability. 
    }
	\label{fig:pn_seq}
\end{figure*}

\begin{figure*}[t]
	\centering
	\subfigure[Mandatory sequence flow.]{\includegraphics[width=0.45\textwidth]{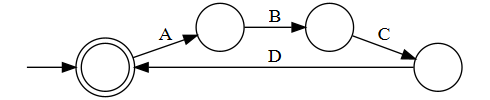} \label{fig:aut_seq_simple}}\hfil
	\subfigure[Optional sequence flow.]{\includegraphics[width=0.45\textwidth]{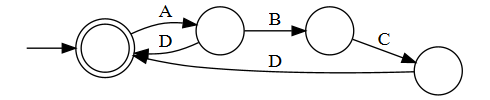}
		\label{fig:aut_seq_optional}}\hfil
	\subfigure[Mandatory repeatable sequence flow.]  {\includegraphics[width=0.45\textwidth]{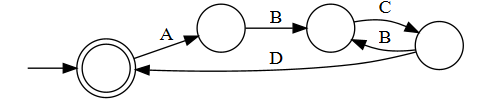}
		\label{fig:aut_seq_repeatable}}\hfil
	\subfigure[Optional repeatable sequence flow.]  {\includegraphics[width=0.45\textwidth]{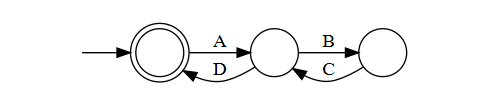}
		\label{fig:aut_seq_optional_repeatable}}
	\caption{Automata of the \declare constraints discovered from the four variants of the sequence flow Petri nets in \cref{fig:pn_seq}.}
	\label{fig:aut_seq}
\end{figure*}

All variants of sequence flow can be detected by relying solely on standard \declare constraints. If there are no repetitions or optional activities within the sequence (cf.~\cref{fig:pn_seq_simple}), it is sufficient to discover the \constraint{Alternate Succession} constraints between the activities. Each activity will have an outgoing \constraint{Alternate Succession} constraint to all subsequent activities, without having any incoming \constraint{Alternate Succession} constraints from preceding activities within the same sequence. By default, this will allow repeating the entire sequence (cf.~\cref{fig:aut_seq_simple}). If such repetition does not occur, then either a \constraint{Not Chain Succession} constraint from the last activity to the first activity in the same sequence will be discovered, or repetition will be implicitly prohibited due to other constraints (e.g., if a single optional activity follows the sequence and has a \constraint{Not Chain Succession} constraint to the first activity of the sequence).

If any optional and/or repeatable activities appear within the sequence, then \constraint{Alternate Succession} constraints will ``jump over'' these activities due to differences in relative activity cardinalities (cf.~\cref{tab:constraint_cardinalities}). For instance, using only \constraint{Alternate Succession} constraints for any of the processes in \cref{fig:pn_seq} (excluding \cref{fig:pn_seq_simple}) would result in two disjoint models: one capturing the sequence \activity{A}, \activity{D}, and another capturing the sequence \activity{B}, \activity{C}.

This issue is resolved for optional sequence flow (cf.~\cref{fig:pn_seq_optional}) by additionally discovering \constraint{Alternate Precedence} and \constraint{Alternate Response} constraints which detect the start and end of the optional region, respectively. Moreover, \constraint{Not Chain Succession} constraints must be discovered to ensure that the sequence \activity{B}, \activity{C} is not repeated. Repeatable sequence flow further requires discovering the \constraint{Succession} constraints; if the sequence is both optional and repeatable (i.e., 0...n iterations), then all \constraint{Precedence} and \constraint{Response} constraints must also be discovered. The corresponding automata representations of the constraints discovered from the Petri nets in \cref{fig:pn_seq} are shown in \cref{fig:aut_seq}.

Mapping for the above sequence flow variants can be derived from the relative activity cardinalities listed in \cref{tab:constraint_cardinalities}, noting that all activities preceding a cardinality change will have corresponding constraints to all activities following that change. For example, in executions of the process in \cref{fig:pn_seq_optional}, the discovered constraints will include both \constraint{Alternate Precedence}(\activity{A}, \activity{B}) and \constraint{Alternate Precedence}(\activity{A}, \activity{C}).

\subsubsection{Parallelism}
\label{sec:parallelism}

\begin{figure*}[t]
    \centering
    \includegraphics[width=0.66\textwidth]{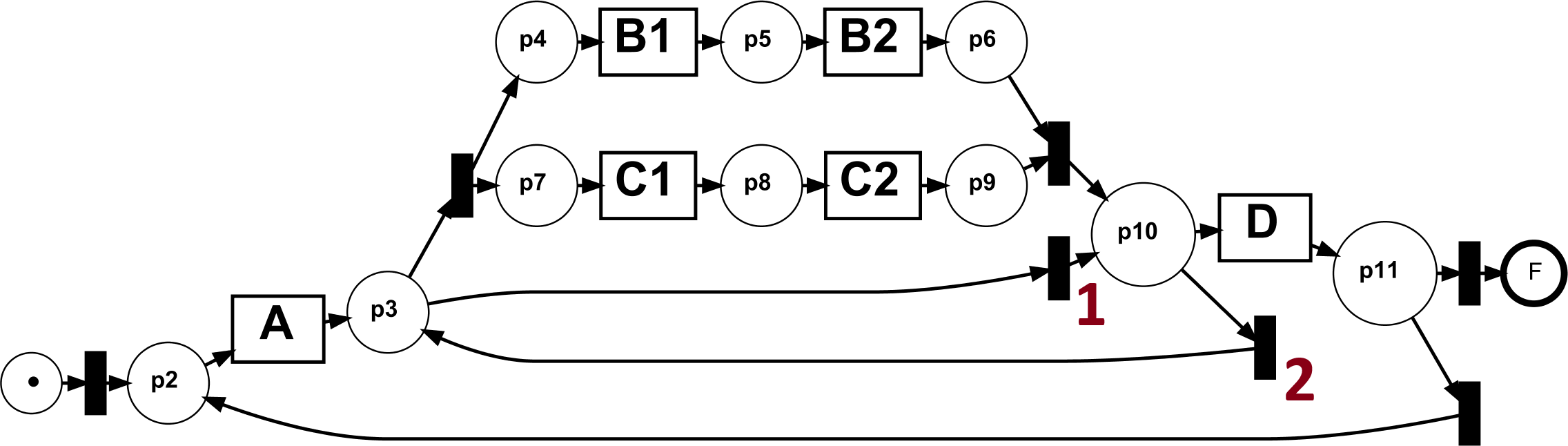}
    \caption{Petri net demonstrating optional repeatable parallelism. Removing the silent transitions marked with 1 and/or 2 yields variants analogous to those in \cref{fig:pn_seq}.
}
    \label{fig:pn_and}
\end{figure*}

\begin{figure*}[t]
	\centering
	\subfigure[Mandatory parallelism.]{\includegraphics[width=0.488\textwidth]{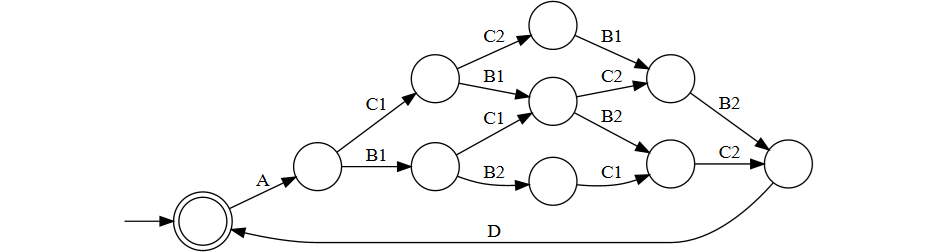} \label{fig:aut_and_simple}}\hfil
	\subfigure[Optional parallelism.]{\includegraphics[width=0.488\textwidth]{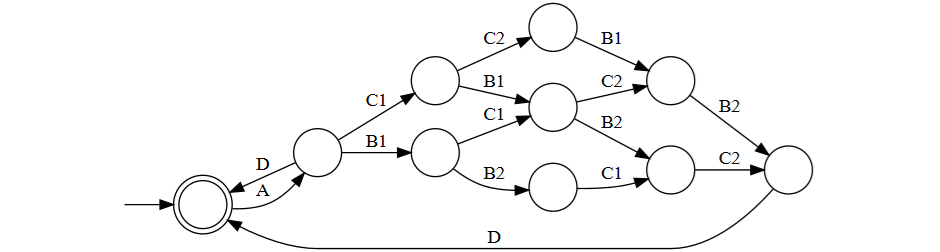}
		\label{fig:aut_and_optional}}\hfil
	\subfigure[Mandatory repeatable parallelism.]  {\includegraphics[width=0.488\textwidth]{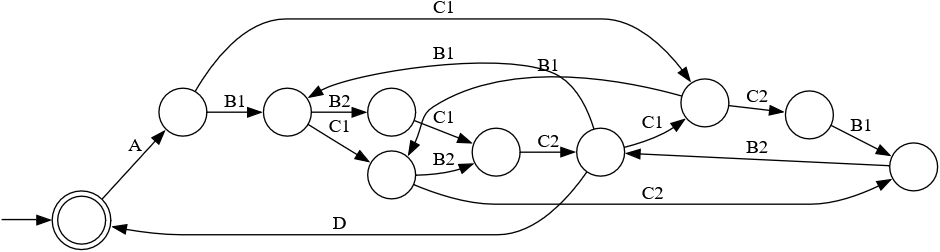}
		\label{fig:aut_and_repeatable}}\hfil
	\subfigure[Optional repeatable parallelism.]  {\includegraphics[width=0.488\textwidth]{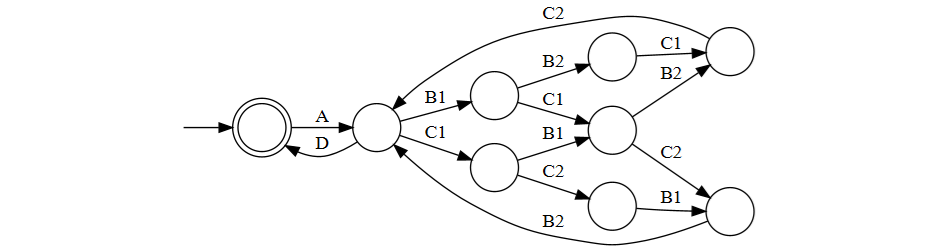}
		\label{fig:aut_and_optional_repeatable}}
	\caption{Automata of the \declare constraints discovered from the four XOR choice variants derived from the Petri net in \cref{fig:pn_and}.
}
	\label{fig:aut_and}
\end{figure*}

The mandatory case of parallelism, where there is exactly one execution of the entire parallel structure with respect to its preceding and following activities, can be obtained from the Petri net in \cref{fig:pn_and} by removing the silent transitions~1 and~2. In this case, the parallel structure can be detected by discovering only \constraint{Alternate Succession} constraints (similar to mandatory sequence flow in \cref{sec:seq_flow}). More specifically, every same-cardinality activity preceding the parallel structure will have an outgoing \constraint{Alternate Succession} constraint to all activities on the parallel branches; every activity on the parallel branches will have an outgoing \constraint{Alternate Succession} constraint to all same-cardinality activities following the parallel structure; and there will be no \constraint{Alternate Succession} constraints between the activities on different parallel branches of the same parallel structure.

\Cref{fig:pn_and} represents the optional repeatable variant of parallelism. The mandatory repeatable and optional variants can be obtained from \cref{fig:pn_and} by removing the silent transitions~1 and~2, respectively. As in \cref{sec:seq_flow}, handling these three variants requires considering the relative activity cardinalities given in \cref{tab:constraint_cardinalities}, and accounting for \constraint{Alternate Succession} constraints ``jumping over'' both optional and repeatable activities.
Furthermore, detecting the correct behavior for these three variants requires discovering the \constraint{Balanced Enablement} constraints, which enforce that all parallel branches must complete before any activity following the parallel join can be executed (including any repetition of the parallel structure), thus eliminating the need for using \constraint{Co-Existence} constraints, which were used in~\cite{DBLP:conf/pmai/AlmanCGMM24}.

As an example, given the process in \cref{fig:pn_and}, the \constraint{Balanced Enablement} constraints would specify that the differences in frequencies among \activity{B1}, \activity{B2}, \activity{C1}, and \activity{C2} must never exceed~1, and that \activity{D} can only occur at a point in the process execution where all aforementioned frequencies are equal. Note that, if the parallel structure must be executed at least once, then \activity{D} will also have an incoming \constraint{Not Chain Succession} from \activity{A}, thus forcing the parallel structure to be executed. The corresponding DFA representations of the discovered constraints are shown in \cref{fig:aut_and_optional,fig:aut_and_repeatable,fig:aut_and_optional_repeatable}.

\subsubsection{Exclusive Choice}
\label{sec:exclusive_choice}

\begin{figure*}[t]
    \centering
    \includegraphics[width=0.58\textwidth]{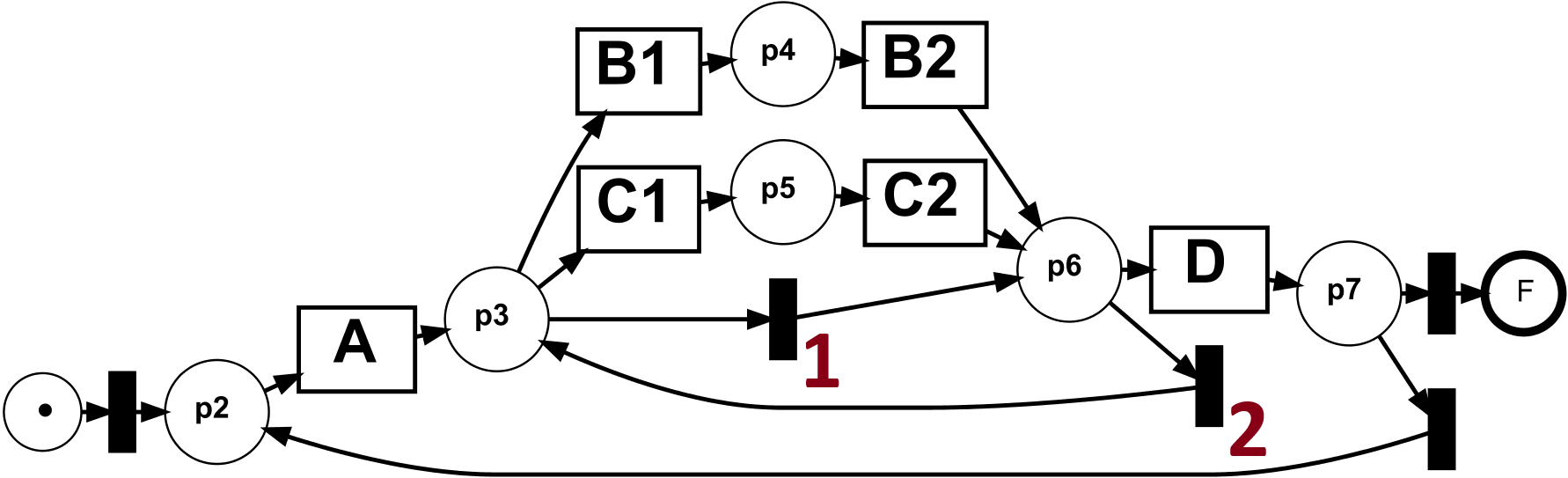}
    \caption{Petri net demonstrating optional repeatable XOR choice. Removing the silent transitions marked with~1 and/or~2 yields variants analogous to those in \cref{fig:pn_seq}.
}
    \label{fig:pn_xor}
\end{figure*}

\begin{figure*}[t]
	\centering
	\subfigure[Mandatory XOR choice.]{\includegraphics[width=0.44\textwidth]{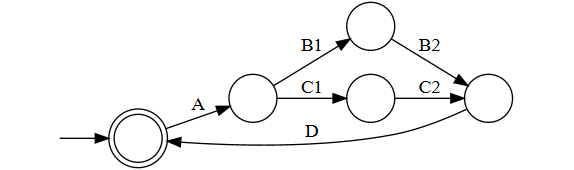} \label{fig:aut_xor_simple}}\hfil
	\subfigure[Optional XOR choice.]{\includegraphics[width=0.44\textwidth]{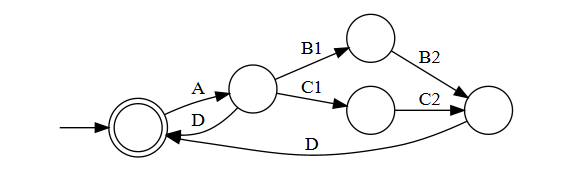}
		\label{fig:aut_xor_optional}}\hfil
	\subfigure[Mandatory repeatable XOR choice.]  {\includegraphics[width=0.44\textwidth]{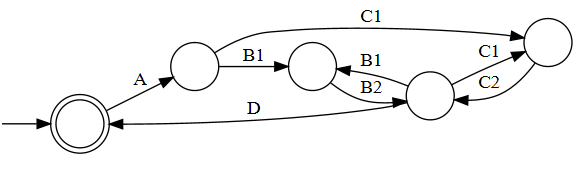}
		\label{fig:aut_xor_repeatable}}\hfil
	\subfigure[Optional repeatable XOR choice.]  {\includegraphics[width=0.44\textwidth]{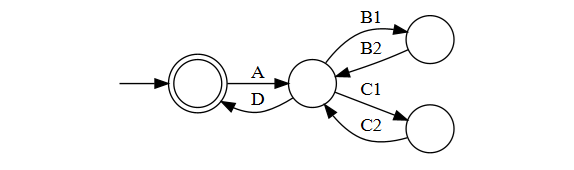}
		\label{fig:aut_xor_optional_repeatable}}
	\caption{Automata of the \declare constraints discovered from the four XOR choice variants derived from the Petri net in \cref{fig:pn_xor}.
}
	\label{fig:aut_xor}
\end{figure*}

Similar to \cref{sec:parallelism}, \cref{fig:pn_xor} shows the optional repeatable variant of exclusive choice (XOR), while the other variants (mandatory, optional, and mandatory repeatable) can be obtained by removing the silent transitions 1, 2, or both. All four variants of XOR can be identified based fully on the standard \declare constraints.

The mandatory and optional XOR variants can be detected using \constraint{Alternate Precedence} and \constraint{Alternate Response}, similar to \constraint{Precedence} and \constraint{Response} in~\cite{DBLP:conf/pmai/AlmanCGMM24}, but also requires discovering \constraint{Not Chain Succession} constraints. For non-repeatable XOR variants, no activity on one XOR branch can be immediately followed by any activity on a different branch of the same XOR construct. The same holds for repeatable XOR variants, with respect to pairs of activities on different branches, except for transitions from the last activities of each branch to the first activities of each branch. 

Both repeatable variants of the XOR construct are detected using \constraint{Precedence} and \constraint{Response} constraints, which will be discovered instead of \constraint{Alternate Precedence} and \constraint{Alternate Response} due to the change in relative cardinalities, as described in \cref{tab:constraint_cardinalities}. Meanwhile, the distinction between non-optional and optional XOR can be determined based on the presence or absence of a \constraint{Not Chain Succession} constraint from the activity directly preceding the XOR to the activity directly following the XOR.

\subsubsection{Inclusive Choice}
\label{sec:inclusive_choice}

\begin{figure*}[t]
    \centering
    \includegraphics[width=0.85\textwidth]{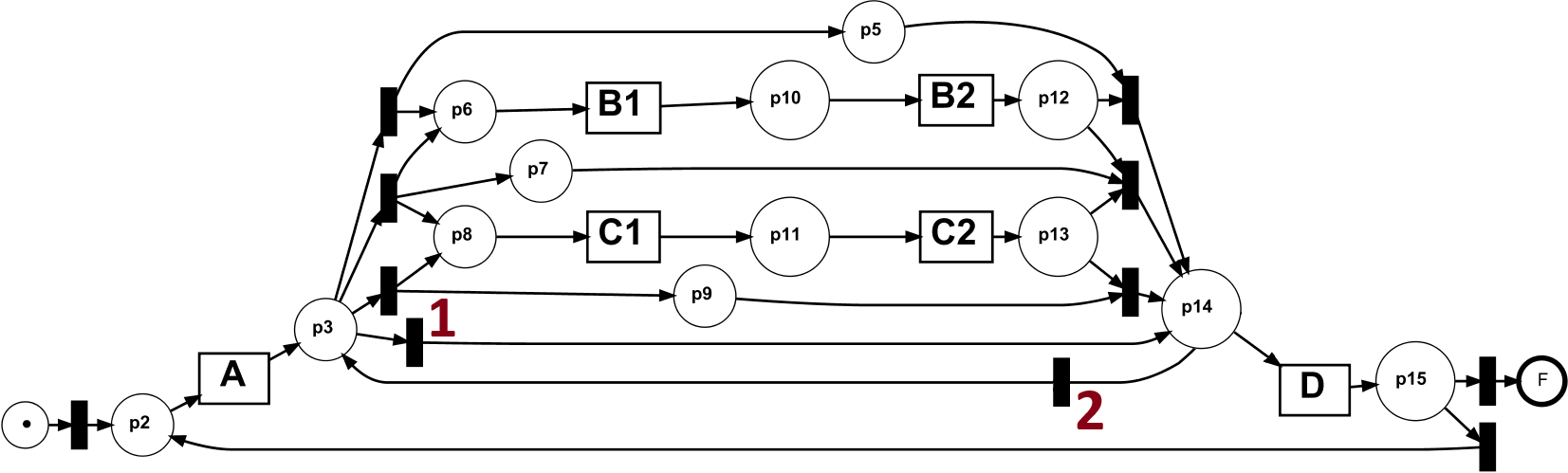}
    \caption{Petri net demonstrating optional repeatable OR choice. Removing the silent transitions marked with 1 and/or 2 yields variants analogous to those in \cref{fig:pn_seq}.
}
    \label{fig:pn_or}
\end{figure*}

\begin{figure*}[t]
	\centering
	\subfigure[Mandatory OR choice.]{\includegraphics[width=0.42\textwidth]{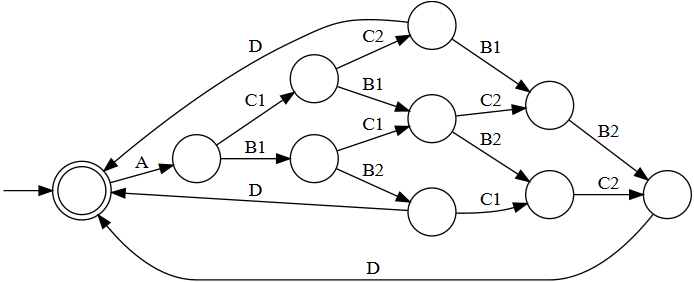} \label{fig:aut_or_simple}}\hfil
	\subfigure[Optional OR choice.]{\includegraphics[width=0.42\textwidth]{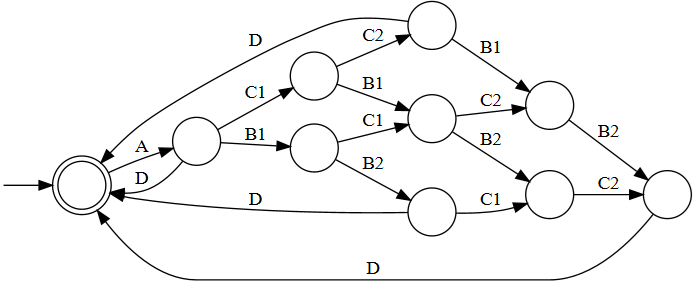}
		\label{fig:aut_or_optional}}\hfil
	\subfigure[Mandatory repeatable OR choice.]  {\includegraphics[width=0.42\textwidth]{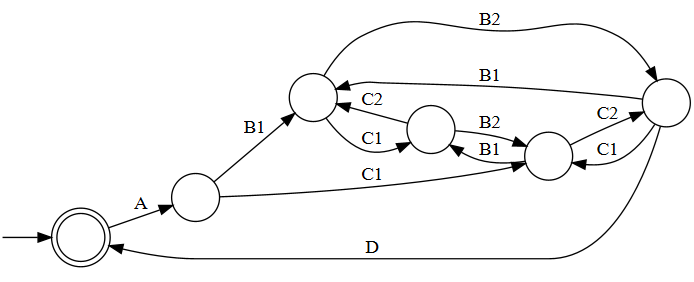}
		\label{fig:aut_or_repeatable}}\hfil
	\subfigure[Optional repeatable OR choice.]  {\includegraphics[width=0.42\textwidth]{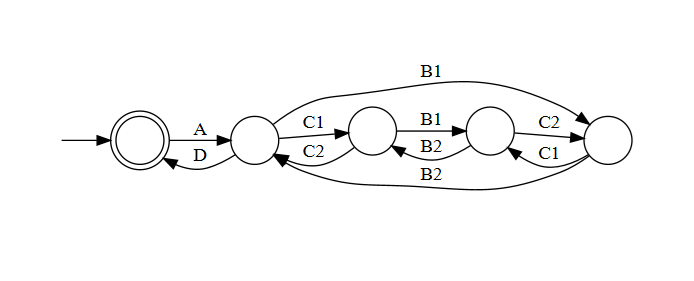}
		\label{fig:aut_or_optional_repeatable}}
	\caption{Automata of the \declare constraints discovered from the four OR choice variants derived from the Petri net in \cref{fig:pn_or}.
}
	\label{fig:aut_or}
\end{figure*}

Similar to \cref{sec:parallelism,sec:exclusive_choice}, \cref{fig:pn_or} shows the optional repeatable variant of inclusive choice (OR), where the mandatory, optional, and mandatory repeatable variants can be obtained by removing the silent transitions 1, 2, or both, respectively. Standard \declare constraints cannot capture any of these variants. The underlying problem is that \declare lacks a mechanisms to track which of the OR branches have been activated (i.e., chosen for execution) and to enforce the completion of the activated branches before any of the activities following the OR construct can be executed (i.e., that all branches chosen for execution complete before the corresponding OR-join occurs).

The \constraint{Interposition} constraint (cf. \cref{sec:constraints}) addresses this limitation for the mandatory and optional OR, as shown in \cref{fig:aut_or_simple} and \cref{fig:aut_or_optional}. For example, given the process in \cref{fig:pn_or}, the discovered constraints would include \constraint{Interposition}(\activity{B1}, \activity{B2}, \activity{D}) and \constraint{Interposition}(\activity{C1}, \activity{C2}, \activity{D}), enforcing the completion of the corresponding OR branches whenever they are selected for execution. However, this is still insufficient for the mandatory repeatable and optional repeatable variants of OR, as illustrated in \cref{fig:aut_or_repeatable} and \cref{fig:aut_or_optional_repeatable}, respectively. In both cases, even if the \constraint{Interposition} constraints are discovered, it remains possible to reach a state in which one branch can be repeated indefinitely while the other branch is not completed. For example, \activity{A}, \activity{B1}, \activity{C1}, \activity{B2} leads to a state where the OR branch containing \activity{B1} and \activity{B2} can be executed infinitely many times before completing the other branch. 

Note that \constraint{Not Chain Succession} does not resolve this issue either, given that there are valid executions, where the last activity of one branch can be directly followed by the first activity of another. The \constraint{Interposition} constraints can resolve this only if the repetition introduced by silent transition 2 in \cref{fig:pn_or} begins with a non-optional activity or a non-optional XOR. The other cases must be validated through post-processing of the event log.

\subsection{Non-Block-Structured Behaviors}
\label{sec:overlapping_behaviors}

The Petri net models in in \cref{sec:isolated_behaviors} follow the same overall structure, where all activities are executed within a larger outer-loop, with each iteration starting with the activity \activity{A} and ending with the activity \activity{D}. The behavior of the remaining two (cf. \cref{fig:pn_seq}) or four (cf. \cref{fig:pn_and,fig:pn_xor,fig:pn_or}) activities is then changed within this loop to match the specific procedural behavior being considered. While providing a relatively controlled way of analyzing the different behavioral variants considered in this paper, it also results in relatively simple block-structured processes. Such processes are generally regarded as easier to handle, which can be seen as a significant limitation of the entire analysis. 

To extend the analysis beyond such block-structured examples, this section illustrates, without claiming exhaustiveness, that the \declare constraints given in \cref{tab:constraint_cardinalities} can also capture non-block-structured procedural behaviors. More specifically, this section presents four examples that contain multiple instances of the previously considered procedural behaviors (e.g., multiple parallel structures), such that these behaviors overlap with each other. As in \cref{sec:isolated_behaviors}, we create a Petri net for each example, generate an event log, discover \declare constraints, and validate the behavioral equivalence of the discovered constraints by constructing the corresponding DFA.

\subsubsection{Optional Behaviors}
\label{sec:seq_opt_overlap}

\begin{figure}[t]
	\centering
	\subfigure[Input Petri net.]{\includegraphics[width=0.48\textwidth]{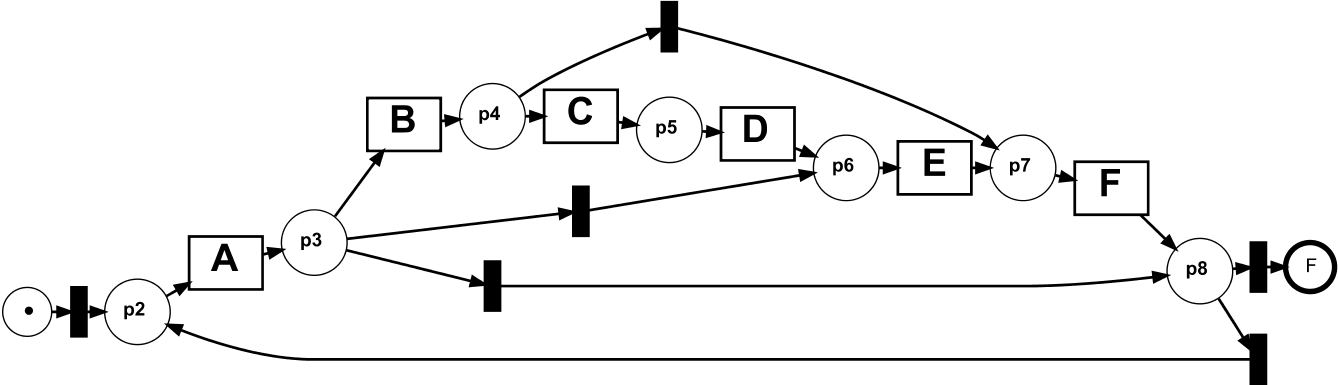} \label{fig:seq_opt_overlap_pn}}\hfil
	\subfigure[DFA of the discovered constraints.]{\includegraphics[width=0.48\textwidth]{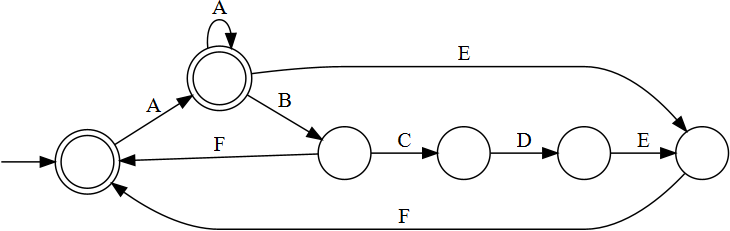}
		\label{fig:seq_opt_overlap_aut}}
	\caption{Petri net example of optional sequence flow with overlapping optional regions, and the corresponding discovered automaton.}
	\label{fig:seq_opt_overlap}
\end{figure}

In this example, we focus on a scenario involving optional sequence flows with overlapping optional regions. In particular, \cref{fig:seq_opt_overlap_pn} presents a Petri net where \activity{A} is mandatory in each iteration of the outer-loop. Following \activity{A}, the execution may either conclude the current iteration, proceed with \activity{B}, or continue with \activity{E}. If the current iteration concludes after \activity{A}, then a subsequent occurrence of \activity{A} may follow immediately after, effectively forming a self-loop that is not explicitly modeled. If \activity{B} is executed after \activity{A}, the process may optionally skip \activity{C}, \activity{D}, and \activity{E}, though \activity{F} remains required. Alternatively, if \activity{E} is executed after \activity{A}, then \activity{F} is again mandatory. As shown in \cref{fig:seq_opt_overlap_aut}, the discovered declarative constraints, inferred from an event log generated from the process in \cref{fig:seq_opt_overlap_pn}, accurately capture this behavior. Notably, the DFA of these constraints explicitly contains the self-loop on \activity{A}.

\subsubsection{Repeatable Behaviors}
\label{sec:seq_rep_overlap}

\begin{figure}[t]
	\centering
	\subfigure[Input Petri net.]{\includegraphics[width=0.48\textwidth]{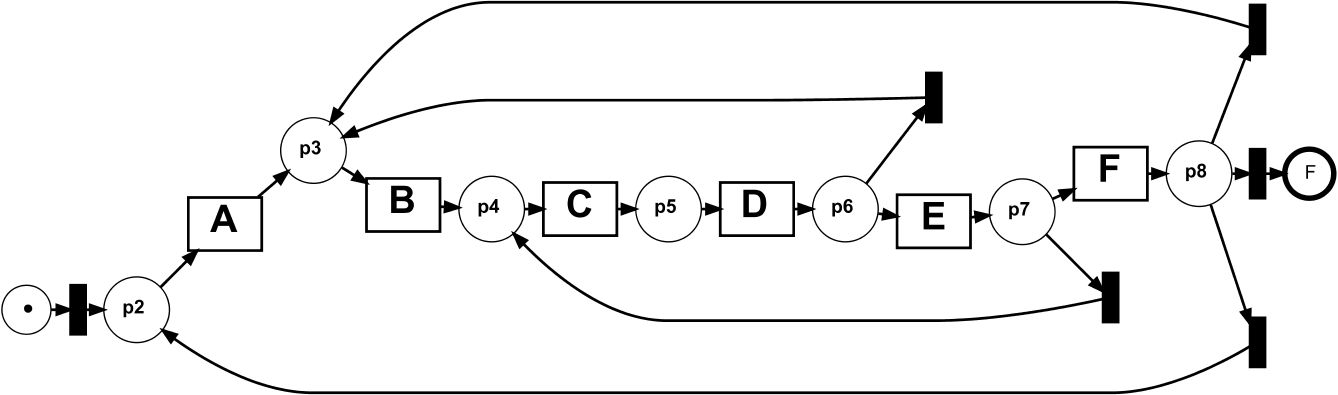} \label{fig:seq_rep_overlap_pn}}\hfil
	\subfigure[DFA of the discovered constraints.]{\includegraphics[width=0.48\textwidth]{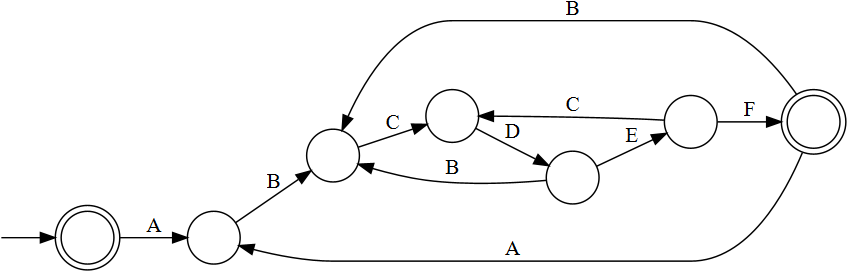}
		\label{fig:seq_rep_overlap_aut}}
	\caption{Petri net example of sequence flow with overlapping repeatable regions, and the corresponding discovered automaton.}
	\label{fig:seq_rep_overlap}
\end{figure}

In this example, we focus on the repeatable sequence flow where some repeatable regions of the process are overlapping. We use a model with the same structure as the one in \cref{sec:seq_opt_overlap}, however, the input and output places of the silent transitions within the outer-loop have been swapped, as shown in \cref{fig:seq_rep_overlap_pn}. This results in a behavior where all activities within the outer-loop body must be executed at least once per iteration of the outer-loop, and there are options to return to an earlier state of the iteration after activities \activity{D}, \activity{E}, and \activity{F}. Notably, the two innermost loops (the sequence \activity{B}, \activity{C}, \activity{D} and the sequence \activity{C}, \activity{D}, \activity{E}) overlap only on the activities \activity{C}, \activity{D}, i.e., neither of these loops is fully contained within the other. Given an event log of the process in \cref{fig:seq_rep_overlap_pn}, the discovered constraints capture this behavior accurately (cf. \cref{fig:seq_rep_overlap_aut}).

\subsubsection{Parallel Branches}
\label{sec:and_overlap}

\begin{figure}[t]
	\centering
	\subfigure[Input Petri net.]{\includegraphics[width=0.85\textwidth]{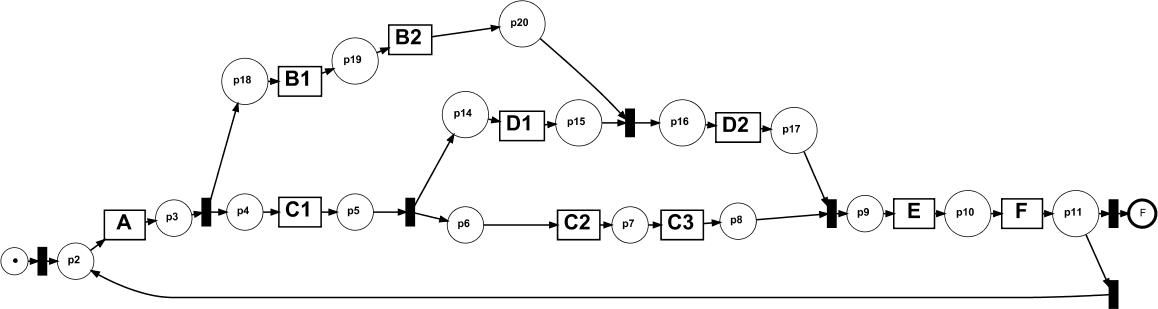} \label{fig:and_overlap_pn}}\hfil
	\subfigure[DFA of the discovered constraints.]{\includegraphics[width=0.85\textwidth]{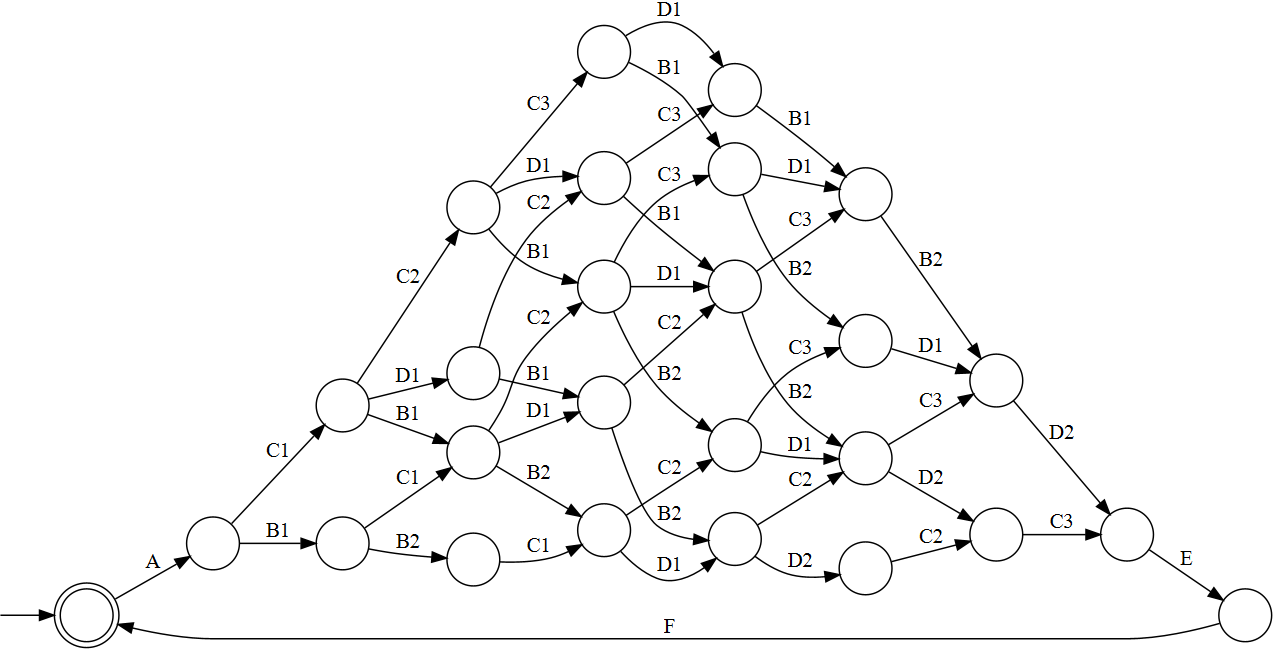}
		\label{fig:and_overlap_aut}}
	\caption{Petri net example of parallelism with overlapping AND splits and joins, and the corresponding discovered automaton.}
	\label{fig:and_overlap}
\end{figure}

In this example, we focus on parallelism, where some parallel splits do not have exact matching parallel joins, i.e., branches of one parallel region are partially overlapping with some parallel branches of a different parallel region. The specific example we consider is given in \cref{fig:and_overlap_pn}. In this example, the branch consisting of \activity{B1}, \activity{B2} merges in the middle of the branch \activity{D1}, \activity{D2}, thus blocking the progression of the latter (the execution of \activity{D2}) until \activity{B2} is executed. Meanwhile, the branch containing \activity{C1}, \activity{C2}, \activity{C3} can progress freely in parallel with \activity{B1}, \activity{B2} as well as with \activity{D1}, \activity{D2}. Notably this Petri net is still deadlock-free and 1-safe. The DFA of the corresponding discovered constraints is shown in \cref{fig:and_overlap_aut}, and it matches the behavior of the Petri net, but through a somewhat less compact structure due to DFAs requiring explicit representation of all possible parallel interleavings.

\subsubsection{Exclusive Choice Branches}
\label{sec:xor_overlap}

\begin{figure}[t]
	\centering
	\subfigure[Input Petri net.]{\includegraphics[width=0.48\textwidth]{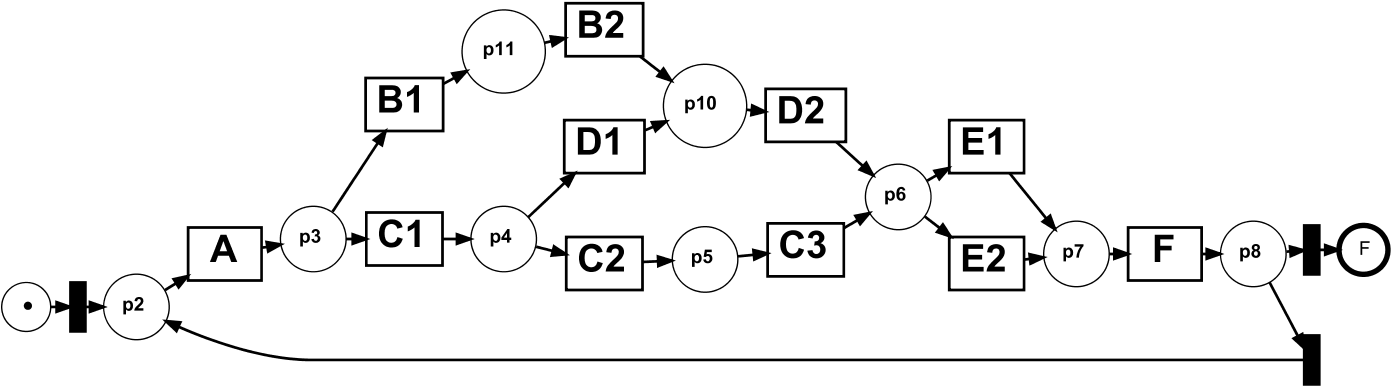} \label{fig:xor_overlap_pn}}\hfil
	\subfigure[DFA of the discovered constraints.]{\includegraphics[width=0.48\textwidth]{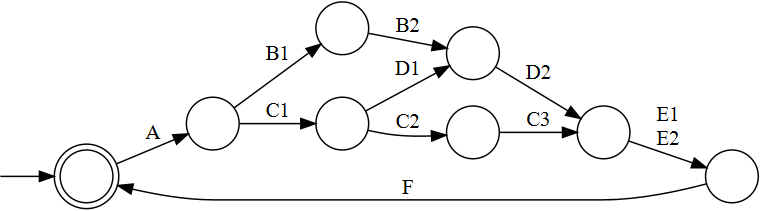}
		\label{fig:xor_overlap_aut}}
	\caption{Petri net example of exclusive choices with overlapping repeatable regions, and the corresponding discovered automaton.}
	\label{fig:xor_overlap}
\end{figure}

In this example, we focus on exclusive choices (XORs), where some XOR-splits do not have exact matching XOR-joins. We use a model with the same structure as the one in \cref{sec:and_overlap}, however, with all parallel splits and joins have been replaced with XOR-splits and XOR-joins, respectively, and \activity{E} has been split into an XOR choice between \activity{E1} and \activity{E2}. This leads to behavior where, for example, choosing \activity{B1} after \activity{A} requires executing \activity{D2}, while choosing \activity{C1} instead of \activity{B1} provides an alternative to \activity{D2} through the choice between \activity{D1} and \activity{C2}. Meanwhile, the split of \activity{E} into \activity{E1} and \activity{E2} is somewhat orthogonal. Instead of demonstrating overlapping behaviors, it demonstrates that consecutive XOR choices (i.e., XOR-join directly followed by a different XOR-split) can also be accurately detected. The DFA of the corresponding discovered constraints is shown in \cref{fig:and_overlap_aut}, and it matches the behavior of the Petri net, while also providing a more compact representation of that behavior.

\section{Related Works}
\label{sec:related}

The ABPMS concept was first introduced in the corresponding manifesto~\cite{DBLP:journals/tmis/DumasFLMMRACGFGRVW23}, with its earliest version appearing online in early 2022.\footnote{arXiv:2201.12855v1 [cs.AI], 30.01.22, URL: \url{https://arxiv.org/abs/2201.12855v1}}
As a result, ABPMS specific research is only starting emerge, with the adoption of large language models (LLMs) being a major focus in, e.g., decision-making through conversational interactions~\cite{DBLP:conf/ital-ia/BernardiCCM24,DBLP:conf/bpm/Casciani24,DBLP:conf/caise/Casciani25,DBLP:conf/rcis/CascianiBCM24}, conversational model redesign~\cite{DBLP:journals/corr/abs-2505-05453}, explainability~\cite{DBLP:journals/corr/abs-2507-23269,DBLP:journals/corr/abs-2504-21032}, and causal reasoning~\cite{DBLP:conf/bpm/FournierLS24}. There are also broader approaches that inject process-awareness into LLMs more directly, such as the Business Process Large Language Model Framework (BPLLM Framework)~\cite{DBLP:journals/jiis/BernardiCCM24}, ProcessLLM~\cite{DBLP:conf/bpm/BussKSW24}, and Large Process Model (LPM)~\cite{DBLP:journals/ki/KampikWRAEGHKHDAPRWW25}.

In addition, multiple other visions, building on the ABPMS concept, have been put forward. For example,~\cite{DBLP:journals/sosym/ChapelaCampaD23} identifies the emergence of augmented process execution, referring to continuous and automated improvement and adaptation of business processes. In the same vein,~\cite{DBLP:conf/pmai/ElyasafMSSAT25} advocates for developing self-modifying autonomous business process systems,~\cite{DBLP:conf/pmai/AcitelliACM24} outlines how an ABPMS can increase business process automation is a real-world business process,~\cite{calvanese2026agenticbusinessprocessmanagement} proposes agentic business process management, while \cite{DBLP:conf/ecai/Montali24} draws a connection between the \declare language and ABPMS.

The above overview is, by no means, comprehensive. However, it illustrates that the use of process modeling languages and the discovery of process models has, thus far, not been explored in light of the broader characteristics of the ABPMS process frame (cf. \cref{sec:frame}). Instead, existing works either use process models as-is, do not use process models at all, or try to embed process knowledge into LLMs directly. These are valid approaches for supporting the development of specific ABPMS related techniques, however, we argue that such approaches will ultimately lead to various ad-hoc representations of process knowledge, each tailored to the specific needs of different techniques.

Our conceptualization of the process frame aims to avoid such ad-hoc representations by providing a language-agnostic generalization that allows existing modeling languages to be combined in a fairly non-restrictive manner. It shares the most similarities with the notion of Hybrid Business Process Representations (HBPRs)~\cite{DBLP:journals/is/AndaloussiBSKW20}, which are also highlighted in the ABPMS manifesto as a potential starting point~\cite{DBLP:journals/tmis/DumasFLMMRACGFGRVW23}. Most HBPRs actually predate the term itself, with one of the earliest examples being ``pockets of flexibility''~\cite{DBLP:conf/er/SadiqSO01,DBLP:journals/is/SadiqOS05}, that allows workflows to contain ``pockets'' where concrete executions are flexibly decided at runtime according to the requirements of that pocket. Interestingly, the original iteration of \declare~\cite{DBLP:conf/edoc/PesicSA07} also supports hierarchical composition of \declare constraints and procedural YAWL workflows~\cite{DBLP:books/sp/yawl10}, making it another early example of an HBPR.

Hierarchical compositions of declarative and procedural models are also explored in multiple other works. For example,~\cite{DBLP:conf/otm/SlaatsSMR16} presents formal semantics based on Petri nets and \declare, where any Petri net transition may require the completion of a \declare sub-model, and any activity of a \declare model may require the completion of a Petri net sub-model. A corresponding process discovery approach, on the basis of Petri nets and \declare, is developed in~\cite{DBLP:conf/bpm/MaggiSR14}. This semantics is general in the sense that other languages, such as BPMN~\cite{BPMN_spec} for procedural models and DCR Graphs~\cite{DBLP:conf/bpm/DeboisHS14,DBLP:journals/corr/abs-1110-4161} for declarative models, could also be integrated by following the same overarching principles. This generality is leveraged, for example, in~\cite{DBLP:conf/bis/SchunselaarSMRA18} to develop a process discovery approach which combines \declare with process trees~\cite{DBLP:conf/apn/LeemansFA13} instead of relying on Petri nets.

Although hierarchy provides a clear separation of declarative and procedural behaviors, it also assumes that these behaviors do not overlap in other ways, i.e., a sub-part of a process can only be declarative or procedural, but not a mix of both. An alternative is to allow (partial) intertwining of procedural and declarative process models by applying both on the same process execution (semi-)concurrently. This approach is taken in~\cite{DBLP:journals/bise/SmedtWVP16}, where declarative constraints are applied on Petri net transitions, such that the firing of any transition is interpreted as an occurrence of the corresponding activity within the context of all constraints. Discovery of such an intertwined HBPR is tackled in \cite{DBLP:journals/dss/SmedtWV15}, while conformance checking is targeted in~\cite{DSCM21}. We consider these works to be the closest to our conceptualization of the process frame, however, they place strong emphasis on single business processes in isolation.

There are also some similarities to the discovery of \emph{local processes}~\cite{DBLP:journals/jides/TaxSHA16}, where the goal is to discover common (and possibly concurrent) sub-behaviors within a single process. Conceptually, this could be reformulated such that the sub-behaviors to be discovered are instead interpreted as parallel processes for which the process identifier is missing. Nonetheless,~\cite{DBLP:journals/jides/TaxSHA16} is not developed with our setting in mind, does not consider declarative behaviors, and, as shown in~\cite{DBLP:journals/topnoc/BruningsFD22}, typically produces very small models, often containing only 2--3 activities, with little to no behavioral overlaps.

In relation to local processes, we highlight region-based discovery techniques, which use region theory to discover characteristic regions that match the behavior observed in the log. These regions, formulated as Petri nets, represent partial views of the behavior captured in the log and are integrated into a complete process model~\cite{carmona2012projection,sole2011region}. While our approach shares certain conceptual similarities with region-based discovery, it relies on \declare discovery and hybrid semantics, offering the flexibility to represent any relation between (or within) processes declaratively, especially when such relations are difficult to encode in Petri nets. There are also conceptual parallels with multi-agent systems (MAS), where larger systems are modeled through decentralized decision units, referred to as agents. Although MAS primarily focuses on the behavioral rules of agents rather than on business processes per se, there is some research exploring the integration of MAS with process discovery techniques, e.g., the work in~\cite{DBLP:conf/edoc/BemthuisKMBIM19}.

A relatively recent research area, that entails a somewhat similar concurrent use of multiple models, is object-centric process mining~\cite{math11122691}. It aims to replace the traditional case-centric view of business processes, where activities are executed within the context of process instances, with an object-centric view, where each activity is executed in the context of one or more instances of (possibly different) types of objects. It that way, it becomes natural to represent a business process as a set of interdependent models, each corresponding to a specific type of object. This is especially apparent in object-centric Petri nets~\cite{DBLP:journals/fuin/AalstB20}, OC-DCR~\cite{DBLP:conf/icpm/ChristfortRFHS24}, and even the object-centric use of event knowledge graphs~\cite{DBLP:books/sp/22/Fahland22}, where activities and control-flow related to different objects can be relatively easily distinguished. However, our conceptualization is not tied to specific objects, but rather, positions each specification as a representation of some logical component of the overall behavioral boundary of the ABPMS. Furthermore, the work presented in this paper is more in line with traditional case-centric view of business processes, in the sense that it assumes only a single overarching case identifier (as opposed to identifiers of multiple different objects) that allows for a meaningful sequential grouping of activity executions.

A potentially viable alternative to the approach considered in this paper could be to employ segmentation techniques, such as the one in~\cite{DBLP:conf/bpm/GuntherRA09}, to determine which activities might belong to which specifications, which would then allow using existing discovery techniques on the resulting activity subsets. However, our strong emphasis on semi-concurrency, activity overlaps, and both declarative and procedural behaviors, is likely to cause issues with adopting these approaches as-is.

Finally, we highlight the work in~\cite{pnToDecl_lexicon}, which provides a comprehensive lexicon of Petri net constructs that accurately capture the meanings of individual \declare constraints. However, most of these Petri net constructs involve the use of weighted, reset, and inhibitor arcs, which are not necessary for the procedural behaviors considered in this paper. Furthermore, this paper does not attempt to transform arbitrary \declare constraints, but rather, focuses on identifying a fixed set of common procedural behaviors based on their manifestation within the \declare language.

\section{Conclusion and Future Work}
\label{sec:conclusion}

In this paper, we presented a conceptualization of the ABPMS process frame as a set of semi-concurrently executed procedural and declarative process specifications, where each specification represents some logical component of the overall ABPMS behavioral boundary, and specifications interact through partial activity overlaps. In addition, we provided a corresponding language-agnostic formalization, based on trace projections, and explored the automated discovery of such a process frame. For the latter, we outlined the rationale for using various eventually-follows relations, instead of directly-follows relations, also for discovering the procedural specifications under the open-world assumption. In particular, we focused on 16 common procedural constructs, from sequence flows to inclusive choices, and demonstrated that 14 of these can be identified from automatically discovered \declare constraints, even in the presence of repetitions in the overall control flow.

The work in this paper allows for numerous future research directions. Focus can be placed on lifting some of the limitations of this paper by, for example, considering additional procedural constructs, removing the implicit use of directly-follows relations (in the form of \constraint{Not Chain Successions}), incorporating other types of eventually-follows relations, and considering interactions of specification beyond the control flow perspective. In addition, this paper involves only Petri nets and \declare, while there are numerous other languages that could be explicitly considered (cf. \cref{sec:conceptualization}), both from representational and from automated process (frame) discovery perspectives.

However, the development of a corresponding complete process discovery approach will most likely be the most impactful and interesting research direction. In particular, process frame, as presented in this paper, allows for multiple behaviorally equivalent representations, where the same behavioral aspects can be expressed, not only using different modeling languages, but also using different modeling paradigms. From the process discovery perspective, it means that a single ``ground truth'' process frame is unlikely to exist for the majority of potential discovery inputs. Instead, factors, such as the ratio of declarative and procedural specifications, the size of individual specifications, the extent of specification overlaps, etc. become relevant, especially in cases where the process frame will also be analyzed (and possibly modified) by human agents. Consequently, automated discovery task should incorporate an element of optimization with respect to (a combination of) the aforementioned factors, which could be tackled, for example, by adopting genetic process discovery algorithms~\cite{DeMedeirosPhD2018}.

\smallskip
\noindent
\textbf{Acknowledgments.}
The work of A.\ Alman was supported by the \mbox{Estonian} Research Council grant PRG1226.

\bibliographystyle{elsarticle-num}
\bibliography{bibliography}

@inproceedings{DBLP:conf/pmai/AlmanCGMM24,
  author       = {Anti Alman and
                  Izack Cohen and
                  Avigdor Gal and
                  Fabrizio Maria Maggi and
                  Marco Montali},
  title        = {Discovering Process Framing for {AI}-Augmented {BPM} Systems in a Multi-Process
                  Setting},
  booktitle    = {PMAI@ECAI},
  series       = {{CEUR} Workshop Proceedings},
  volume       = {3779},
  pages        = {47--58},
  publisher    = {CEUR-WS.org},
  year         = {2024}
}

@article{ALMAN2023102512,
    title = {Monitoring hybrid process specifications with conflict management: An automata-theoretic approach},
    journal = {Artificial Intelligence in Medicine},
    volume = {139},
    pages = {102512},
    year = {2023},
    author = {Anti Alman and Fabrizio Maria Maggi and Marco Montali and Fabio Patrizi and Andrey Rivkin}
}

@article{ALMAN2023102271,
    title = {A framework for modeling, executing, and monitoring hybrid multi-process specifications with bounded global–local memory},
    journal = {Information Systems},
    volume = {119},
    pages = {102271},
    year = {2023},
    author = {Anti Alman and Fabrizio Maria Maggi and Marco Montali and Fabio Patrizi and Andrey Rivkin}
}

@inproceedings{Maggi2026,
author="Maggi, Fabrizio Maria
and Alman, Anti
and Wittlinger, Paul Hermann",
title="From Constraint-Based Process Modeling to Framed Autonomy: A Historical Excursus",
bookTitle="Mining a Scientist's Process: Essays Dedicated to Wil van der Aalst on the Occasion of His 60th Birthday",
year="2026",
publisher="Springer Nature Switzerland",
address="Cham",
pages="199--211"
}

@inproceedings{DBLP:conf/bpm/WittlingerAAMM25,
  author       = {Paul Hermann Wittlinger and
                  Giacomo Acitelli and
                  Anti Alman and
                  Fabrizio Maria Maggi and
                  Andrea Marrella},
  title        = {FrAIm: {A} What-If Analysis Tool Enabling Framed Autonomy via Automated
                  Planning},
  booktitle    = {{BPM} (Demos / Resources Forum)},
  series       = {{CEUR} Workshop Proceedings},
  pages        = {192--199},
  publisher    = {CEUR-WS.org},
  year         = {2025}
}

@article{DBLP:journals/tmis/DumasFLMMRACGFGRVW23,
  author       = {Marlon Dumas and
                  Fabiana Fournier and
                  Lior Limonad and
                  Andrea Marrella and
                  Marco Montali and
                  Jana{-}Rebecca Rehse and
                  Rafael Accorsi and
                  Diego Calvanese and
                  Giuseppe {De Giacomo} and
                  Dirk Fahland and
                  Avigdor Gal and
                  Marcello {La Rosa} and
                  Hagen V{\"{o}}lzer and
                  Ingo Weber},
  title        = {{AI}-augmented Business Process Management Systems: {A} Research Manifesto},
  journal      = {{ACM} Trans. Manag. Inf. Syst.},
  volume       = {14},
  number       = {1},
  pages        = {11:1--11:19},
  year         = {2023}
}

@misc{calvanese2026agenticbusinessprocessmanagement,
      title={Agentic Business Process Management: A Research Manifesto}, 
      author={Diego Calvanese and Angelo Casciani and Giuseppe De Giacomo and Marlon Dumas and Fabiana Fournier and Timotheus Kampik and Emanuele La Malfa and Lior Limonad and Andrea Marrella and Andreas Metzger and Marco Montali and Daniel Amyot and Peter Fettke and Artem Polyvyanyy and Stefanie Rinderle-Ma and Sebastian Sardiña and Niek Tax and Barbara Weber},
      year={2026},
      eprint={2603.18916},
      archivePrefix={arXiv},
      primaryClass={cs.AI}
}

@inproceedings{DBLP:conf/edoc/PesicSA07,
  author       = {Maja Pesic and
                  Helen Schonenberg and
                  Wil M. P. van der Aalst},
  title        = {{DECLARE:} {F}ull Support for Loosely-Structured Processes},
  booktitle    = {{EDOC}},
  pages        = {287--300},
  publisher    = {{IEEE} Computer Society},
  year         = {2007}
}

@inproceedings{DBLP:conf/icse/DwyerAC99,
  author       = {Matthew B. Dwyer and
                  George S. Avrunin and
                  James C. Corbett},
  title        = {Patterns in Property Specifications for Finite-State Verification},
  booktitle    = {{ICSE}},
  pages        = {411--420},
  publisher    = {{ACM}},
  year         = {1999}
}

@book{DBLP:series/lnbip/Montali10,
  author       = {Marco Montali},
  title        = {Specification and Verification of Declarative Open Interaction Models
                  - {A} Logic-Based Approach},
  series       = {Lecture Notes in Business Information Processing},
  volume       = {56},
  publisher    = {Springer},
  year         = {2010}
}

@article{DBLP:journals/pieee/Murata89,
  author       = {Tadao Murata},
  title        = {Petri nets: Properties, analysis and applications},
  journal      = {Proc. {IEEE}},
  volume       = {77},
  number       = {4},
  pages        = {541--580},
  year         = {1989}
}

@article{DBLP:journals/topnoc/HeeSW13a,
  author       = {Kees M. {van Hee} and
                  Natalia Sidorova and
                  Jan Martijn E. M. van der Werf},
  title        = {Business Process Modeling Using {Petri} Nets},
  journal      = {Trans. Petri Nets Other Model. Concurr.},
  volume       = {7},
  pages        = {116--161},
  year         = {2013}
}

@article{DBLP:journals/is/MaggiCFK18,
  author       = {Fabrizio Maria Maggi and
                  Claudio {Di Ciccio} and
                  Chiara {Di Francescomarino} and
                  Taavi Kala},
  title        = {Parallel algorithms for the automated discovery of declarative process
                  models},
  journal      = {Inf. Syst.},
  volume       = {74},
  number       = {Part},
  pages        = {136--152},
  year         = {2018}
}

@inproceedings{DBLP:conf/caise/MaggiBA13,
  author       = {Fabrizio Maria Maggi and
                  R. P. Jagadeesh Chandra Bose and
                  Wil M. P. van der Aalst},
  title        = {A Knowledge-Based Integrated Approach for Discovering and Repairing
                  Declare Maps},
  booktitle    = {CAiSE},
  series       = {Lecture Notes in Computer Science},
  volume       = {7908},
  pages        = {433--448},
  publisher    = {Springer},
  year         = {2013}
}

@inproceedings{DBLP:conf/bpm/LeemansFA13,
  author       = {Sander J. J. Leemans and
                  Dirk Fahland and
                  Wil M. P. van der Aalst},
  title        = {Discovering Block-Structured Process Models from Event Logs Containing
                  Infrequent Behaviour},
  booktitle    = {Business Process Management Workshops},
  series       = {Lecture Notes in Business Information Processing},
  pages        = {66--78},
  publisher    = {Springer},
  year         = {2013}
}

@inproceedings{DBLP:conf/er/SadiqSO01,
  author       = {Shazia Sadiq and
                  Wasim Sadiq and
                  Maria E. Orlowska},
  title        = {Pockets of Flexibility in Workflow Specification},
  booktitle    = {{ER}},
  series       = {Lecture Notes in Computer Science},
  pages        = {513--526},
  publisher    = {Springer},
  year         = {2001}
}

@inproceedings{DBLP:conf/bpm/AlmanMMR23,
  author       = {Anti Alman and
                  Fabrizio Maria Maggi and
                  Marco Montali and
                  Andrey Rivkin},
  title        = {Generating Event Logs from Hybrid Process Models},
  booktitle    = {Business Process Management Workshops},
  series       = {Lecture Notes in Business Information Processing},
  volume       = {492},
  pages        = {289--301},
  publisher    = {Springer},
  year         = {2023}
}

@article{DBLP:journals/icae/WeijtersA03,
  author       = {A. J. M. M. Weijters and
                  Wil M. P. van der Aalst},
  title        = {Rediscovering workflow models from event-based data using little thumb},
  journal      = {Integr. Comput. Aided Eng.},
  volume       = {10},
  number       = {2},
  pages        = {151--162},
  year         = {2003}
}

@article{DBLP:journals/kais/AugustoCDRP19,
  author       = {Adriano Augusto and
                  Raffaele Conforti and
                  Marlon Dumas and
                  Marcello {La Rosa} and
                  Artem Polyvyanyy},
  title        = {Split miner: Automated discovery of accurate and simple business process
                  models from event logs},
  journal      = {Knowl. Inf. Syst.},
  volume       = {59},
  number       = {2},
  pages        = {251--284},
  year         = {2019}
}

@techreport{pnToDecl_lexicon,
author = {De Smedt, Johannes and {vanden Broucke}, Seppe and Weerdt, Jochen and Vanthienen, Jan},
year = {2015},
month = {02},
pages = {},
title = {A Full {R/I}-net Construct Lexicon for Declare Constraints},
institution = {KU Leuven},
journal = {SSRN Electronic Journal}
}

@inproceedings{DBLP:conf/bpm/BandaraLRM21,
  author       = {Wasana Bandara and
                  Amy {Van Looy} and
                  Michael Rosemann and
                  Lara Meyers},
  title        = {A call for 'Holistic' Business Process Management},
  booktitle    = {Problems@BPM},
  series       = {{CEUR} Workshop Proceedings},
  volume       = {2938},
  pages        = {6--10},
  publisher    = {CEUR-WS.org},
  year         = {2021}
}

@article{DBLP:journals/sttt/KupfermanV03,
  author       = {Orna Kupferman and
                  Moshe Y. Vardi},
  title        = {Vacuity detection in temporal model checking},
  journal      = {Int. J. Softw. Tools Technol. Transf.},
  volume       = {4},
  number       = {2},
  pages        = {224--233},
  year         = {2003}
}

@article{DBLP:journals/is/SmedtWSV18,
  author       = {Johannes {De Smedt} and
                  Jochen {De Weerdt} and
                  Estefan{\'{\i}}a Serral and
                  Jan Vanthienen},
  title        = {Discovering hidden dependencies in constraint-based declarative process
                  models for improving understandability},
  journal      = {Inf. Syst.},
  volume       = {74},
  number       = {Part},
  pages        = {40--52},
  year         = {2018}
}

@inproceedings{DBLP:conf/bpm/AlmanCMMA21,
  author       = {Anti Alman and
                  Claudio {Di Ciccio} and
                  Fabrizio Maria Maggi and
                  Marco Montali and
                  Han van der Aa},
  title        = {{RuM}: Declarative Process Mining, Distilled},
  booktitle    = {{BPM}},
  series       = {Lecture Notes in Computer Science},
  volume       = {12875},
  pages        = {23--29},
  publisher    = {Springer},
  year         = {2021}
}

@article{DBLP:journals/is/AcitelliAMM25,
  author       = {Giacomo Acitelli and
                  Anti Alman and
                  Fabrizio Maria Maggi and
                  Andrea Marrella},
  title        = {Achieving framed autonomy in {AI}-augmented business process management
                  systems through automated planning},
  journal      = {Inf. Syst.},
  volume       = {133},
  pages        = {102573},
  year         = {2025}
}

@inproceedings{DBLP:conf/caise/AlmanMRRW24,
  author       = {Anti Alman and
                  Fabrizio Maria Maggi and
                  Stefanie Rinderle{-}Ma and
                  Andrey Rivkin and
                  Karolin Winter},
  title        = {Towards a Multi-model Paradigm for Business Process Management},
  booktitle    = {CAiSE},
  series       = {Lecture Notes in Computer Science},
  volume       = {14663},
  pages        = {178--194},
  publisher    = {Springer},
  year         = {2024}
}

@article{DBLP:journals/tmis/CiccioM15,
  author       = {Claudio Di Ciccio and
                  Massimo Mecella},
  title        = {On the Discovery of Declarative Control Flows for Artful Processes},
  journal      = {{ACM} Trans. Manag. Inf. Syst.},
  volume       = {5},
  number       = {4},
  pages        = {24:1--24:37},
  year         = {2015}
}

@article{carmona2012projection,
  title={Projection approaches to process mining using region-based techniques},
  author={Carmona, Josep},
  journal={Data Mining and Knowledge Discovery},
  volume={24},
  pages={218--246},
  year={2012},
  publisher={Springer}
}

@article{sole2011region,
  title={Region-based foldings in process discovery},
  author={Sol{\'e}, Marc and Carmona, Josep},
  journal={IEEE Transactions on Knowledge and Data Engineering},
  volume={25},
  number={1},
  pages={192--205},
  year={2011},
  publisher={IEEE}
}

@article{DBLP:journals/jides/TaxSHA16,
  author       = {Niek Tax and
                  Natalia Sidorova and
                  Reinder Haakma and
                  Wil M. P. van der Aalst},
  title        = {Mining local process models},
  journal      = {J. Innov. Digit. Ecosyst.},
  volume       = {3},
  number       = {2},
  pages        = {183--196},
  year         = {2016}
}

@article{DBLP:journals/topnoc/BruningsFD22,
  author       = {Mitchel Brunings and
                  Dirk Fahland and
                  Boudewijn F. van Dongen},
  title        = {Defining Meaningful Local Process Models},
  journal      = {Trans. Petri Nets Other Model. Concurr.},
  volume       = {16},
  pages        = {24--48},
  year         = {2022}
}

@article{DBLP:journals/dss/SmedtWV15,
  author    = {Johannes {De Smedt} and
               Jochen {De Weerdt} and
               Jan Vanthienen},
  title     = {Fusion Miner: Process discovery for mixed-paradigm models},
  journal   = {Decis. Support Syst.},
  volume    = {77},
  pages     = {123--136},
  year      = {2015}
}

@inproceedings{DBLP:conf/bpm/MaggiSR14,
  author    = {Fabrizio Maria Maggi and
               Tijs Slaats and
               Hajo A. Reijers},
  title     = {The Automated Discovery of Hybrid Processes},
  booktitle = {{BPM}},
  series    = {Lecture Notes in Computer Science},
  volume    = {8659},
  pages     = {392--399},
  publisher = {Springer},
  year      = {2014}
}

@article{DBLP:journals/is/AndaloussiBSKW20,
  author    = {Amine Abbad Andaloussi and
               Andrea Burattin and
               Tijs Slaats and
               Ekkart Kindler and
               Barbara Weber},
  title     = {On the declarative paradigm in hybrid business process representations:
               {A} conceptual framework and a systematic literature study},
  journal   = {Inf. Syst.},
  volume    = {91},
  pages     = {101505},
  year      = {2020}
}

@inproceedings{DBLP:conf/edoc/BemthuisKMBIM19,
  author       = {Rob H. Bemthuis and
                  Martijn Koot and
                  Martijn R. K. Mes and
                  Faiza Allah Bukhsh and
                  Maria{-}Eugenia Iacob and
                  Nirvana Meratnia},
  title        = {An Agent-Based Process Mining Architecture for Emergent Behavior Analysis},
  booktitle    = {{EDOC} Workshops},
  pages        = {54--64},
  publisher    = {{IEEE}},
  year         = {2019}
}

@TechReport{BPMN_spec,
	author = {OMG},
	title = {Business Process Model and Notation ({BPMN})},
    institution = {Object Management Group},
	type = {Technical Report},
	year = {2014},
    url = {https://www.omg.org/spec/BPMN/2.0.2/PDF},
    note = {Version 2.0.2}
}

@inproceedings{DBLP:conf/bpm/DeboisHS14,
  author       = {S{\o}ren Debois and
                  Thomas T. Hildebrandt and
                  Tijs Slaats},
  title        = {Hierarchical Declarative Modelling with Refinement and Sub-processes},
  booktitle    = {{BPM}},
  volume       = {8659},
  pages        = {18--33},
  publisher    = {Springer},
  year         = {2014}
}

@inproceedings{DBLP:journals/corr/abs-1110-4161,
  author       = {Thomas T. Hildebrandt and
                  Raghava Rao Mukkamala},
  title        = {Declarative Event-Based Workflow as Distributed Dynamic Condition
                  Response Graphs},
  booktitle    = {{PLACES}},
  series       = {{EPTCS}},
  volume       = {69},
  pages        = {59--73},
  year         = {2010}
}

@TechReport{SBVR_spec,
	author = {OMG},
	title = {Semantics Of Business Vocabulary And Business Rules {(SBVR)}},
    institution = {Object Management Group},
	type = {Technical Report},
	year = {2019},
    url = {https://www.omg.org/spec/SBVR/1.5/},
    note = {(accessed 22.01.26)}
}

@inproceedings{DBLP:conf/semweb/BoleyTW01,
  author       = {Harold Boley and
                  Said Tabet and
                  Gerd Wagner},
  title        = {Design Rationale for {RuleML}: {A} Markup Language for Semantic Web
                  Rules},
  booktitle    = {{SWWS}},
  pages        = {381--401},
  year         = {2001}
}

@article{mcguinness2004owl,
  title={{OWL} web ontology language overview},
  author={McGuinness, Deborah L and Van Harmelen, Frank},
  journal={W3C recommendation},
  year={2004},
  url = {http://www.w3.org/TR/2004/REC-owl-features-20040210/},
  note = {(accessed 22.01.26)}
}

@inproceedings{DBLP:conf/iceis/BenevidesG09,
  author       = {Alessander Botti Benevides and
                  Giancarlo Guizzardi},
  title        = {A Model-Based Tool for Conceptual Modeling and Domain Ontology Engineering
                  in OntoUML},
  booktitle    = {{ICEIS}},
  series       = {Lecture Notes in Business Information Processing},
  volume       = {24},
  pages        = {528--538},
  publisher    = {Springer},
  year         = {2009}
}

@book{sterling2009art,
  title={The art of agent-oriented modeling},
  author={Sterling, Leon and Taveter, Kuldar},
  year={2009},
  publisher    = {MIT Press}

}

@phdthesis{thesis_Pesic,
title = "Constraint-based workflow management systems : shifting control to users",
author = "M. Pesic",
year = "2008",
type = "{PhD} Thesis",
school = "Technische Universiteit Eindhoven"
}

@inproceedings{DBLP:conf/bpm/BarWL25,
  author       = {Philipp B{\"{a}}r and
                  Moe Thandar Wynn and
                  Sander J. J. Leemans},
  title        = {A Full Picture in Conformance Checking: Efficiently Summarizing All
                  Optimal Alignments},
  booktitle    = {{BPM}},
  series       = {Lecture Notes in Computer Science},
  pages        = {69--87},
  publisher    = {Springer},
  year         = {2025}
}

@inproceedings{DBLP:conf/bpm/KustersA25,
  author       = {Aaron K{\"{u}}sters and
                  Wil M. P. van der Aalst},
  title        = {{OC-DECLARE}: Discovering Object-Centric Declarative Patterns with
                  Synchronization},
  booktitle    = {{BPM}},
  series       = {Lecture Notes in Computer Science},
  pages        = {162--179},
  publisher    = {Springer},
  year         = {2025}
}

@article{DBLP:journals/fuin/AalstB20,
  author       = {Wil M. P. van der Aalst and
                  Alessandro Berti},
  title        = {Discovering Object-centric {Petri} Nets},
  journal      = {Fundam. Informaticae},
  volume       = {175},
  number       = {1-4},
  pages        = {1--40},
  year         = {2020}
}

@inproceedings{DBLP:conf/sefm/Aalst19,
  author       = {Wil M. P. van der Aalst},
  title        = {Object-Centric Process Mining: Dealing with Divergence and Convergence
                  in Event Data},
  booktitle    = {{SEFM}},
  series       = {Lecture Notes in Computer Science},
  pages        = {3--25},
  publisher    = {Springer},
  year         = {2019}
}

@inproceedings{DBLP:conf/icpm/DettenSL24,
  author       = {Jan Niklas van Detten and
                  Pol Schumacher and
                  Sander J. J. Leemans},
  title        = {Discovering Compact, Live and Identifier-Sound Object-Centric Process
                  Models},
  booktitle    = {{ICPM}},
  pages        = {113--120},
  publisher    = {{IEEE}},
  year         = {2024}
}

@inproceedings{DBLP:conf/caise/HaisjacklZSHRPW13,
  author       = {Cornelia Haisjackl and
                  Stefan Zugal and
                  Pnina Soffer and
                  Irit Hadar and
                  Manfred Reichert and
                  Jakob Pinggera and
                  Barbara Weber},
  title        = {Making Sense of Declarative Process Models: Common Strategies and
                  Typical Pitfalls},
  booktitle    = {{BMMDS/EMMSAD}},
  series       = {Lecture Notes in Business Information Processing},
  pages        = {2--17},
  publisher    = {Springer},
  year         = {2013}
}

@inproceedings{DBLP:conf/bpm/GuntherRA09,
  author       = {Christian W. G{\"{u}}nther and
                  Anne Rozinat and
                  Wil M. P. van der Aalst},
  title        = {Activity Mining by Global Trace Segmentation},
  booktitle    = {Business Process Management Workshops},
  series       = {Lecture Notes in Business Information Processing},
  pages        = {128--139},
  publisher    = {Springer},
  year         = {2009}
}

@inproceedings{DBLP:conf/bpm/Westergaard11,
  author       = {Michael Westergaard},
  title        = {Better Algorithms for Analyzing and Enacting Declarative Workflow
                  Languages Using {LTL}},
  booktitle    = {{BPM}},
  series       = {LNCS},
  volume       = {6896},
  pages        = {83--98},
  publisher    = {Springer},
  year         = {2011}
}

@book{DBLP:books/daglib/0027363,
  author       = {Wil M. P. van der Aalst},
  title        = {Process Mining - Discovery, Conformance and Enhancement of Business
                  Processes},
  publisher    = {Springer},
  year         = {2011}
}

@inproceedings{DBLP:conf/ital-ia/BernardiCCM24,
  author       = {Mario Luca Bernardi and
                  Angelo Casciani and
                  Marta Cimitile and
                  Andrea Marrella},
  title        = {A preliminary study on Business Process-aware Large Language Models},
  booktitle    = {Ital-IA},
  series       = {{CEUR} Workshop Proceedings},
  volume       = {3762},
  pages        = {441--446},
  publisher    = {CEUR-WS.org},
  year         = {2024}
}

@inproceedings{DBLP:conf/bpm/Casciani24,
  author       = {Angelo Casciani},
  title        = {Conversational {AI} for Framed Autonomy in {AI}-augmented Business Process
                  Management},
  booktitle    = {{BPM} (Demos / Resources Forum)},
  series       = {{CEUR} Workshop Proceedings},
  volume       = {3758},
  pages        = {53--60},
  publisher    = {CEUR-WS.org},
  year         = {2024}
}

@inproceedings{DBLP:conf/caise/Casciani25,
  author       = {Angelo Casciani},
  title        = {Integrating {LLMs} and Symbolic Reasoning for Framed Autonomy in {AI}-Augmented
                  Business Process Management},
  booktitle    = {CAiSE Forum},
  series       = {Lecture Notes in Business Information Processing},
  volume       = {557},
  pages        = {277--285},
  publisher    = {Springer},
  year         = {2025}
}

@inproceedings{DBLP:conf/rcis/CascianiBCM24,
  author       = {Angelo Casciani and
                  Mario Luca Bernardi and
                  Marta Cimitile and
                  Andrea Marrella},
  title        = {Conversational Systems for {AI}-Augmented Business Process Management},
  booktitle    = {{RCIS} {(1)}},
  series       = {Lecture Notes in Business Information Processing},
  volume       = {513},
  pages        = {183--200},
  publisher    = {Springer},
  year         = {2024}
}

@article{DBLP:journals/corr/abs-2505-05453,
  author       = {Nataliia Klievtsova and
                  Timotheus Kampik and
                  Juergen Mangler and
                  Stefanie Rinderle{-}Ma},
  title        = {Conversational Process Model Redesign},
  journal      = {CoRR},
  volume       = {abs/2505.05453},
  year         = {2025}
}

@article{DBLP:journals/corr/abs-2507-23269,
  author       = {Peter Fettke and
                  Fabiana Fournier and
                  Lior Limonad and
                  Andreas Metzger and
                  Stefanie Rinderle{-}Ma and
                  Barbara Weber},
  title        = {{XABPs}: Towards eXplainable Autonomous Business Processes},
  journal      = {CoRR},
  volume       = {abs/2507.23269},
  year         = {2025}
}

@article{DBLP:journals/corr/abs-2504-21032,
  author       = {Lior Limonad and
                  Fabiana Fournier and
                  Hadar Mulian and
                  George Manias and
                  Spiros Borotis and
                  Danai Kyrkou},
  title        = {Selecting the Right {LLM} for {eGov} Explanations},
  journal      = {CoRR},
  volume       = {abs/2504.21032},
  year         = {2025}
}

@inproceedings{DBLP:conf/bpm/FournierLS24,
  author       = {Fabiana Fournier and
                  Lior Limonad and
                  Inna Skarbovsky},
  title        = {Towards a Benchmark for Causal Business Process Reasoning with {LLMs}},
  booktitle    = {Business Process Management Workshops},
  series       = {Lecture Notes in Business Information Processing},
  volume       = {534},
  pages        = {233--246},
  publisher    = {Springer},
  year         = {2024}
}

@article{DBLP:journals/jiis/BernardiCCM24,
  author       = {Mario Luca Bernardi and
                  Angelo Casciani and
                  Marta Cimitile and
                  Andrea Marrella},
  title        = {Conversing with business process-aware large language models: the
                  {BPLLM} framework},
  journal      = {J. Intell. Inf. Syst.},
  volume       = {62},
  number       = {6},
  pages        = {1607--1629},
  year         = {2024}
}

@inproceedings{DBLP:conf/bpm/BussKSW24,
  author       = {Alina Buss and
                  Wolfgang Kratsch and
                  Sebastian Johannes Schmid and
                  Hongyang Wang},
  title        = {{ProcessLLM}: {A} Large Language Model Specialized in the Interpretation,
                  Analysis, and Optimization of Business Processes},
  booktitle    = {Business Process Management Workshops},
  series       = {Lecture Notes in Business Information Processing},
  volume       = {534},
  pages        = {221--232},
  publisher    = {Springer},
  year         = {2024}
}

@article{DBLP:journals/ki/KampikWRAEGHKHDAPRWW25,
  author       = {Timotheus Kampik and
                  Christian Warmuth and
                  Adrian Rebmann and
                  Ron Agam and
                  Lukas N. P. Egger and
                  Andreas Gerber and
                  Johannes Hoffart and
                  Jonas Kolk and
                  Philipp Herzig and
                  Gero Decker and
                  Han van der Aa and
                  Artem Polyvyanyy and
                  Stefanie Rinderle{-}Ma and
                  Ingo Weber and
                  Matthias Weidlich},
  title        = {Large Process Models: {A} Vision for Business Process Management in
                  the Age of Generative {AI}},
  journal      = {K{\"{u}}nstliche Intell.},
  volume       = {39},
  number       = {2},
  pages        = {81--95},
  year         = {2025}
}

@article{DBLP:journals/sosym/ChapelaCampaD23,
  author       = {David Chapela{-}Campa and
                  Marlon Dumas},
  title        = {From process mining to augmented process execution},
  journal      = {Softw. Syst. Model.},
  volume       = {22},
  number       = {6},
  pages        = {1977--1986},
  year         = {2023}
}

@inproceedings{DBLP:conf/pmai/ElyasafMSSAT25,
  author       = {Achiya Elyasaf and
                  Andreas Metzger and
                  Sebastian Sardina and
                  Arik Senderovich and
                  Estefan{\'{\i}}a Serral Asensio and
                  Niek Tax},
  title        = {Toward Self-Modifying Autonomous Business Process Systems},
  booktitle    = {{PMAI}},
  series       = {{CEUR} Workshop Proceedings},
  publisher    = {CEUR-WS.org},
  year         = {2025}
}

@inproceedings{DBLP:conf/ecai/Montali24,
  author       = {Marco Montali},
  title        = {{AI} for Declarative Processes: Representation, Mining, Synthesis},
  booktitle    = {{ECAI}},
  series       = {Frontiers in Artificial Intelligence and Applications},
  volume       = {392},
  pages        = {17--24},
  publisher    = {{IOS} Press},
  year         = {2024}
}

@inproceedings{DBLP:conf/pmai/AcitelliACM24,
  author       = {Giacomo Acitelli and
                  Simone Agostinelli and
                  Angelo Casciani and
                  Andrea Marrella},
  title        = {An End-To-End Execution of a Logistic Process in an {AI}-Augmented Business
                  Process Management System},
  booktitle    = {PMAI@ECAI},
  series       = {{CEUR} Workshop Proceedings},
  volume       = {3779},
  pages        = {5--10},
  publisher    = {CEUR-WS.org},
  year         = {2024}
}

@article{DBLP:journals/is/SadiqOS05,
  author    = {Shazia Wasim Sadiq and
               Maria E. Orlowska and
               Wasim Sadiq},
  title     = {Specification and validation of process constraints for flexible workflows},
  journal   = {Inf. Syst.},
  volume    = {30},
  number    = {5},
  pages     = {349--378},
  year      = {2005}
}

@book{DBLP:books/sp/yawl10,
  editor       = {Arthur H. M. ter Hofstede and
                  Wil M. P. van der Aalst and
                  Michael Adams and
                  Nick Russell},
  title        = {Modern Business Process Automation - {YAWL} and its Support Environment},
  publisher    = {Springer},
  year         = {2010}
}

@inproceedings{DBLP:conf/otm/SlaatsSMR16,
  author       = {Tijs Slaats and
                  Dennis M. M. Schunselaar and
                  Fabrizio Maria Maggi and
                  Hajo A. Reijers},
  title        = {The Semantics of Hybrid Process Models},
  booktitle    = {{OTM} Conferences},
  volume       = {10033},
  pages        = {531--551},
  year         = {2016}
}

@inproceedings{DBLP:conf/apn/LeemansFA13,
  author       = {Sander J. J. Leemans and
                  Dirk Fahland and
                  Wil M. P. van der Aalst},
  title        = {Discovering Block-Structured Process Models from Event Logs - {A}
                  Constructive Approach},
  booktitle    = {Petri Nets},
  series       = {Lecture Notes in Computer Science},
  volume       = {7927},
  pages        = {311--329},
  publisher    = {Springer},
  year         = {2013}
}

@inproceedings{DBLP:conf/bis/SchunselaarSMRA18,
  author       = {Dennis M. M. Schunselaar and
                  Tijs Slaats and
                  Fabrizio Maria Maggi and
                  Hajo A. Reijers and
                  Wil M. P. van der Aalst},
  title        = {Mining Hybrid Business Process Models: {A} Quest for Better Precision},
  booktitle    = {{BIS}},
  series       = {Lecture Notes in Business Information Processing},
  volume       = {320},
  pages        = {190--205},
  publisher    = {Springer},
  year         = {2018}
}

@article{DBLP:journals/bise/SmedtWVP16,
  author       = {Johannes {De Smedt} and
                  Jochen {De Weerdt} and
                  Jan Vanthienen and
                  Geert Poels},
  title        = {Mixed-Paradigm Process Modeling with Intertwined State Spaces},
  journal      = {Bus. Inf. Syst. Eng.},
  volume       = {58},
  number       = {1},
  pages        = {19--29},
  year         = {2016}
}

@article{DSCM21,
  author    = {Boudewijn F. van Dongen and
               Johannes {De Smedt} and
               Claudio {Di Ciccio} and
               Jan Mendling},
  title     = {Conformance checking of mixed-paradigm process models},
  journal   = {Inf. Syst.},
  volume    = {102},
  year      = {2021}
}

@inproceedings{DBLP:conf/icpm/ChristfortRFHS24,
  author       = {Axel Kjeld Fjelrad Christfort and
                  Andrey Rivkin and
                  Dirk Fahland and
                  Thomas T. Hildebrandt and
                  Tijs Slaats},
  title        = {Discovery of Object-Centric Declarative Models},
  booktitle    = {{ICPM}},
  pages        = {121--128},
  publisher    = {{IEEE}},
  year         = {2024}
}

@incollection{DBLP:books/sp/22/Fahland22,
  author       = {Dirk Fahland},
  title        = {Process Mining over Multiple Behavioral Dimensions with Event Knowledge
                  Graphs},
  booktitle    = {Process Mining Handbook},
  series       = {Lecture Notes in Business Information Processing},
  volume       = {448},
  pages        = {274--319},
  publisher    = {Springer},
  year         = {2022}
}

@article{math11122691,
AUTHOR = {van der Aalst, Wil M. P.},
TITLE = {Object-Centric Process Mining: Unraveling the Fabric of Real Processes},
JOURNAL = {Mathematics},
VOLUME = {11},
YEAR = {2023},
NUMBER = {12},
ARTICLE-NUMBER = {2691}
}

@phdthesis{DeMedeirosPhD2018,
  author  = {De Medeiros, A. K. Alves},
  title   = {Genetic process mining},
  school  = {Technical University of Eindhoven},
  year    = {2006}
}

\end{document}